\documentclass[lettersize,journal]{IEEEtran}
\usepackage{amsmath,amsfonts}
\usepackage{algorithmic}
\usepackage{algorithm}
\usepackage{array}
\usepackage[caption=false,font=normalsize,labelfont=sf,textfont=sf]{subfig}
\usepackage{textcomp}
\usepackage{stfloats}
\usepackage{url}
\usepackage{verbatim}
\usepackage{graphicx}
\usepackage{cite}
\usepackage{bm}
\usepackage{multirow}
\usepackage{booktabs}
\usepackage{graphicx}
\usepackage{xcolor}

\long\def\comment#1{}

\usepackage{hyperref}  

\usepackage{xcolor}
\usepackage{color, colortbl}
\usepackage{xr-hyper}

\def\red#1{\textcolor{red}{#1}}

\begin{document}

\title{FauForensics: Boosting Audio-Visual Deepfake Detection with Facial Action Units}

\author{
    Jian Wang, 
    Baoyuan Wu,~\IEEEmembership{Senior Member,~IEEE,} 
    Li Liu,~\IEEEmembership{Senior Member,~IEEE,}
    Qingshan Liu,~\IEEEmembership{Senior Member,~IEEE}
\thanks{Jian Wang and Qingshan Liu are with the School of Computer Science, Nanjing University of Posts and Telecommunications, Nanjing 210046, PR China. E-mail: \{jwang.cs, qsliu\}@njupt.edu.cn. (Corresponding author: Qingshan Liu.)}
\thanks{Baoyuan Wu is with School of Data Science, The Chinese University of Hong Kong, Shenzhen, Guangdong, 518172, P.R. China (E-mail: wubaoyuan@cuhk.edu.cn).}
\thanks{Li Liu is with The Hong Kong University of Science and Technology (Guangzhou), Guangdong, 511458, China (E-mail: avrillliu@hkust-gz.edu.cn).}

}

\markboth{Journal of \LaTeX\ Class Files,~Vol.~14, No.~8, August~2021}%
{Shell \MakeLowercase{\textit{et al.}}: A Sample Article Using IEEEtran.cls for IEEE Journals}


\maketitle

\begin{abstract}

    The rapid evolution of generative AI has intensified the threat of realistic audio-visual deepfakes, demanding robust and generalizable detection methods. Existing solutions primarily address unimodal (\textit{e.g.}, audio, visual) forgeries but struggle with multimodal manipulations due to inadequate handling of heterogeneous modality features and poor cross-dataset generalization. We propose \textbf{FauForensics}, a novel framework leveraging biologically invariant facial action units (FAUs), which are quantitative descriptors of facial muscle activity linked to emotion physiology. They serve as forgery-resistant representations that reduce domain dependency while capturing subtle synthetic-content disruptions. In addition, unlike prior clip-level comparisons, our method computes frame-wise audio-visual similarities via a fusion module with learnable cross-modal queries, dynamically aligning lip-audio relationships and mitigating feature heterogeneity. Experiments on \textbf{FakeAVCeleb} and \textbf{LAV-DF} show state-of-the-art performance with 4.83\% average cross-dataset improvement over existing methods.

\end{abstract}

\begin{IEEEkeywords}
    Deepfake Detection, Multimodal, Facial Action Units, Implicit Feature Alignment, Temporal Correlation.
\end{IEEEkeywords}

\section{Introduction}
\label{sec:intro}

    Compared with unimodal deepfakes, advanced multimodal forgeries can bring a more realistic experience, which raises increasing security concerns about multimodal deepfakes in the community. Recently, researchers have proposed some audio-visual deepfake detectors to capture multimodal forgery traces and obtain promising performance. However, increasingly advanced and various multimodal forgery attacks bring new challenges and demands to current detectors, especially accurately and simultaneously capturing various forgeries, including audio forgery, visual forgery, and multimodal forgery. Therefore, it is urgent to develop a powerful and robust multi-task deepfake detector against various audio-visual forgery attacks.

\begin{figure}[t]
  \centering
   \includegraphics[width=1\linewidth]{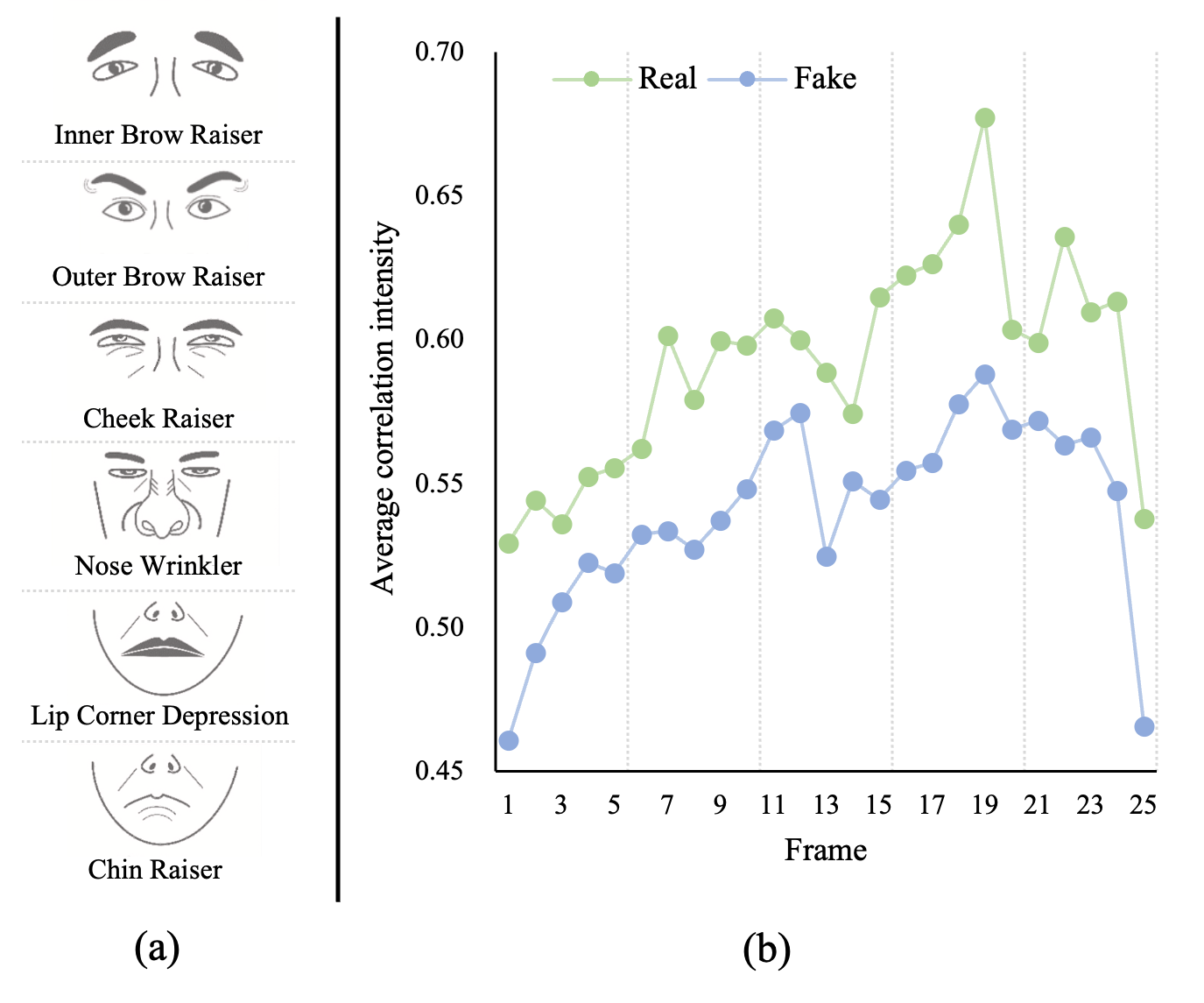}
   \caption{(a) Some examples of Facial Action Units~\cite{blasberg2023you}. FAUs depict the movement of specific facial muscles. (b) The average correlation intensity analysis of FAUs across the temporal domain between real and fake videos is conducted on over 20,000 samples. Compared with real videos, fake videos show a weaker temporal correlation intensity.}
   \label{fig:story}
\end{figure}

    Previous audio-visual deepfake detectors usually focus on inter-modal dissonance~\cite{chugh2020not}, ensemble learning~\cite{hashmi2022multimodal,hashmi2025avtenet}, and multimodal joint learning~\cite{yu2023pvass} to capture audio-visual forgery traces. For example, Chugh~\emph{et al.}~\cite{chugh2020not} computed modality dissonance score between audio and visual features via a contrastive learning method for detecting multimodal forgery, which lacks inter-modal feature interaction. Hashmi~\emph{et al.}~\cite{hashmi2022multimodal} utilize an ensemble-based method to distinguish the authenticity of the audio-video pair. In summary, these methods either lack inter-modal or intra-modal interaction, which limits the detector's ability to comprehensively capture various forgeries, including (1) audio-only forgeries (e.g., synthetic voice generation, audio splicing), (2) visual-only forgeries (e.g., face swap), and (3) multi-modal forgeries (e.g., misaligned lip-sync, mismatched audio-visual events). Subsequently, some approaches~\cite{liu2023mcl,zou2024cross,oorloff2024avff} attempt to focus on both inter- and intra-modal interaction, but these methods are only used for detecting multi-modal forgeries and not for unimodal forgeries. Recently, a few work~\cite{zhang2024joint,yin2024fine} have paid attention to this problem and proposed methods to simultaneously capture all three types of forgeries. Although these methods have gained promising performance under within-database settings, most of them easily suffer from obvious performance degradation when handling unseen audio-visual signals.

    To address these challenges, it is essential to develop a generalized and robust detection framework capable of effectively capturing intrinsic unimodal/multimodal forgery traces for accurate detection. One promising solution is to transfer intrinsic knowledge learned from real audio-visual signals via large-scale multimodal pretraining to downstream detection tasks. For example, Oorloff~\emph{et al.}~\cite{oorloff2024avff} propose a two-stage cross-modal learning method to capture the intrinsic audio-visual correspondences via self-supervision on large-scale real videos. However, these pre-training methods rely on multi-stage training with large-scale datasets are time-consuming and resource-hungry. Furthermore, cross-modal biological invariance between modalities presents another possible solution for multimodal forgery detection. For example, early work~\cite{mittal2020emotions} utilizes emotion consistency to capture affective forgery cues between audio-visual signals. However, emotion-only forgery detectors can not handle increasingly improved multimodal deepfakes. Therefore, it is essential to capture more effective physiologically constrained relationships between modalities, as these are inherently difficult to perfectly replicate in synthetic multimedia content.

    In this work, we present FauForensics, an audio-visual deepfake detection framework guided by Facial Action Units (FAUs) to identify subtle manipulation patterns and enhance generalization. Our key insight stems from analyzing the temporal correlation intensity of FAUs: real videos exhibit significantly higher consistency in FAU-based muscle movements compared to forged ones (Fig.~\ref{fig:story}(b)), where discontinuous facial dynamics reveal cross-modal dissonance. The framework operates through four stages (see Fig. \ref{fig:main}):
    \textbf{FAU-enhanced feature learning}: A frozen pre-trained FAU encoder extracts fine-grained micro-motion features, while a trainable video encoder captures global visual context.
    \textbf{Implicit feature alignment}: A query-shared multimodal transformer implicitly aligns audio-visual representations via attention mechanisms.
    \textbf{Temporal attentional pooler}: We introduce a novel frame-wise temporal attentional pooling module to jointly capture intra-modality inconsistencies (e.g., irregular eye blinks) and inter-modality conflicts (e.g., mismatched lip-audio sync).
    \textbf{Multiple forgery detection}: Pooled features are processed through modality-specific MLPs to predict authenticity scores for unimodal/multimodal scenarios.
    This design shifts focus from clip-level coarse comparisons to frame-level spatiotemporal anomaly detection driven by physiologically grounded FAUs, significantly improving robustness against evolving forgery techniques. 

In summary, the main contributions of this work are three-fold:
\begin{itemize}
   \item We propose a novel FAU-guided framework integrating frame-level audio-visual fusion and temporal consistency modeling to detect unseen deepfakes. Particularly, FAUs enhance visual forgery localization by capturing micro-expression inconsistencies and inter-modal temporal anomalies simultaneously.
   \item We introduce a temporal attention-based pooling module following implicit cross-modal feature alignment, which dynamically captures intra-/inter-modality temporal dependencies to improve detection accuracy in both unimodal and multimodal deepfake scenarios. 
   \item Experiments on FakeAVCeleb and LAV-DF show state-of-the-art (SOTA) performance. Importantly, it demonstrates superior cross-dataset generalizability with up to an average of 4.83\% to existing methods.
 \end{itemize}

    The rest of the paper is organized as follows: Sec.~\ref{section:rw} reviews related work of Audio-Visual Deepfake Detection, Facial Attribution-Guided Forgery Detection, and Temporal Forgery Detection. The proposed FauForensics architecture is detailed in Sec.~\ref{sec:Method}. Experimental setup, results, and analysis are described in Sec.~\ref{section:exp}. Finally, the paper is concluded in Sec.~\ref{section:conc}.

%
\section{Related Work}\label{section:rw}
    In this section, we give a brief review of Audio-Visual Deepfake Detection, Facial Attribution-Guided Forgery Detection, and Temporal Forgery Detection.
\subsection{Audio-Visual Deepfake Detection}
    Most of the prior work focused on unimodal deepfake detection~\cite{li2020face,kong2022detect,wang2022lisiam,yang2023masked,yan2023ucf,zhang2024robust}, especially visual-only deepfake detection. For example, Yang~\emph{et al.}~\cite{yang2020preventing} proposed a lip-based speaker authentication system that focuses on dynamic lip movement to defend against human imposters. With the rapid development of multimodal generative models~\cite{mukhopadhyay2024diff2lip,li2024latentsync,jiang2024loopy}, some researchers~\cite{zhou2021joint,hashmi2022multimodal,shahzad2022lip,knafo2022fakeout,cozzolino2023audio,ilyas2023avfakenet,raza2023multimodaltrace} were turning to detect multimodal forgery traces using audio-visual signals. For example, Agarwal~\emph{et al.}~\cite{agarwal2020detecting} claimed that the inconsistency between the dynamics of the mouth shape and the spoken phoneme could be used to capture cross-modal manipulation traces. Although they achieved promising performance using hand-crafted features to capture audio-visual forgery traces, they overlook high-level semantic information and the interaction between audio-visual features. To further mine complementary information between modalities, Zhou~\emph{et al.}~\cite{zhou2021joint} proposed a joint detection framework to learn robust feature representations from audio-visual signals. In addition, Zou~\emph{et al.}~\cite{zou2024cross} employed modality-specific and cross-modality regularization to improve multimodal representation learning for audio-visual deepfake detection. These methods inspire us to detect subtle cross-modal manipulation traces using multi-modal signals for audio-visual deepfake detection. Compared with unimodal deepfake detection, multimodal deepfake detection approaches require more inter-modal correlation learning to capture complex dependencies across different modalities. Therefore, in this work, we focus on audio-visual deepfake detection and propose a novel framework to model inter- and intra-modal relationships between audio and visual signals.

\subsection{Facial Attribution-Guided Forgery Detection}
    To better capture audio-visual forgery traces, some researchers attempted to capture multimodal forgery traces from basic facial characteristics. For example, Mittal~\emph{et al.}~\cite{mittal2020emotions} mined emotion cues from the two modalities and detected artifacts by emotion similarity (or dissimilarity) between audio-visual signals. Although these prior works obtained promising performance in within-database evaluations, they easily suffer from performance drop and do not cope with the increasingly advanced and various multimodal forgery attacks. To avoid these problems, recent work~\cite{hal2022l,yu2023pvass,zhao2024audio,oorloff2024avff} attempted to employ self-supervised learning methods pre-trained in a large-scale authentic database to mine facial-related inherent properties, such as identity consistency, lip sync, etc. However, these large-scale pre-training methods are time-consuming and resource-hungry. Recently, Liu~\emph{et al.}~\cite{liu2024lips} exploited the biologically intrinsic correlation between lip movements and head postures to detect lip-syncing deepfakes. Peng~\emph{et al.}~\cite{peng2024deepfakes} proposed a gaze analysis model to capture distribution differences of gaze direction pattern between the real and forged videos. In addition, some visual forgery detectors~\cite{chu2022protecting,bai2023aunet} utilized facial action units (FAUs) to capture visual forgery traces related to facial muscle but neglected to exploit the correlation between audio-visual modalities. Facial Action Units (FAUs) exhibit strong correlations with audio signals through both physiological relationships (e.g., lower-face AUs drive different sound generation) and affective synchronization (e.g., eyebrow movements are correlated with voice frequency)~\cite{meng2017listen}. Therefore, we introduce FAUs to obtain structured geometric information for better capturing audio-visual dissonance between facial muscle movements and audio signals.

\begin{figure*}[ht]
    \centering
    \includegraphics[width=1.0\linewidth]{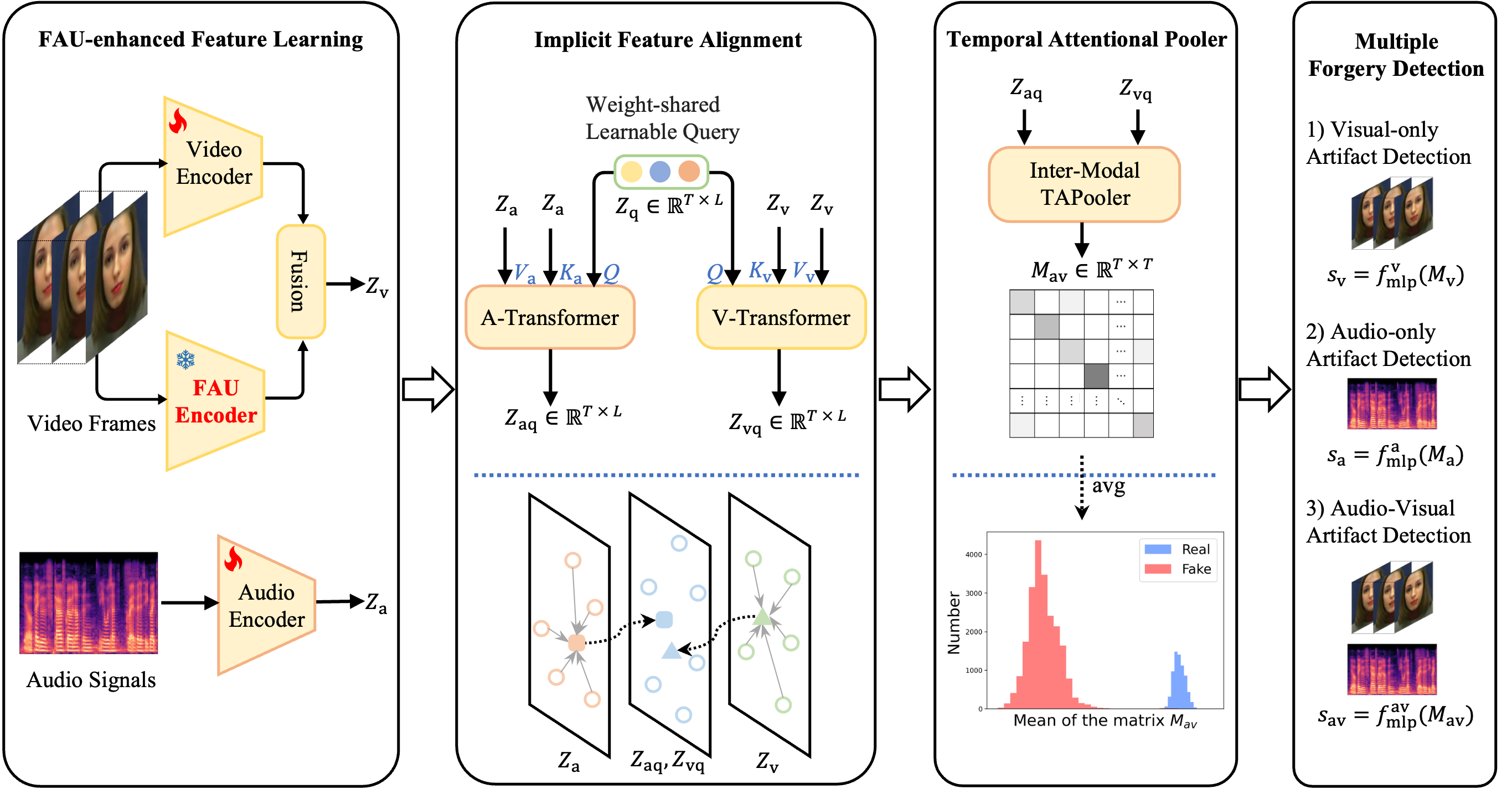}
    \caption{ Overview of the proposed FauForensics framework. Firstly, we take audio-video pairs as inputs and encode them into latent features using two audio/video encoders with learnable weights, alongside a pre-trained FAU encoder with frozen weights. The FAU features are expected to enable detectors to capture FAU-related forgeries. We then perform implicit feature alignment before inter-modal modeling to facilitate more effective audio-visual feature interaction. Next, we design a novel temporal attentional pooler to model intra- and inter-modal temporal correlation for capturing various forgeries. Finally, we feed pooled features into MLP for classification.}
    \label{fig:main}
\end{figure*}

\subsection{Temporal Forgery Detection}
    Previously, some work~\cite{h2021lips,gu2021spatiotemporal,zhao2023istvt,coccomini2024mintime} attempted to capture temporal forgery cues for unimodal deepfake detection. However, these methods rely on intra-modal modeling of unimodality in the temporal domain and ignore inter-modal temporal forgery traces for multimodal deepfake detection. To capture inter-modality temporal forgery traces, most research efforts were devoted to exploring audio-visual dissonance~\cite{gu2021deepfake,feng2023self,cheng2023voice,liu2024lips,bohacek2024lost} and joint audio-visual feature learning~\cite{yang2023avoid,oorloff2024avff,liangspeechforensics,chen2025glcf} for improving generalization across unseen manipulation methods. For example, Zhou~\emph{et al.}~\cite{zhou2021joint} proposed a multimodal joint learning approach to learn robust feature representations from audio-visual modality. Liu~\emph{et al.}~\cite{liu2023mcl} proposed a contrastive learning method to reduce the cross-modal gap and learn a compositional embedding from multimodal signals for capturing audio-visual temporal forgeries. However, these methods focused on clip-level or patch-level semantic interaction between two modalities and ignored key temporal correlations for audio-visual temporal forgery detection. To avoid these issues, a few researchers~\cite{yin2024fine} began to model finer-grained inter-modality relationships via frame-level feature interaction to capture audio-visual dissonance. For example, Yin~\emph{et al.}~\cite{yin2024fine} proposed a graph attention network to learn audio-visual synchronization patterns for capturing audio-visual forgery traces. However, this method relied on randomly sampled frames from video clips to learn clip-level feature interaction between two modalities, which ignored finer-grained frame-by-frame forgery clues. In this work, we explore intra- and inter-modality temporal correlation in a frame-by-frame manner to capture slight temporal artifacts for audio-visual temporal forgery detection.

\section{Method}    \label{sec:Method}

    In this section, we first introduce FauForensics architecture in Sec.\ref{subsection:arch} and then present FAU-enhanced Feature Learning, Implicit Feature Alignment, Temporal Attentional Pooler, and Multiple Forgery Detection in Sec. \ref{subsection:ffl}, Sec. \ref{subsection:ifa}, Sec.\ref{subsection:tap}, and Sec.\ref{subsection:mfd}, respectively. Finally, we describe our implementation details in Sec.\ref{subsection:id}.

\subsection{Architecture}\label{subsection:arch}
  Our proposed FauForensics is a novel audio-visual deepfake detection architecture that aims to learn robust and fine-grained temporal consistency between audio-visual signals. Our architecture consists of FAU-enhanced Feature Learning, Implicit Feature Alignment, Temporal Attentional Pooler, and Multiple Forgery Detection. We depict an overview of our proposed approach in Fig.~\ref{fig:main}. We can see that three different encoders (i.e., $f^{\rm a}_{\rm enc}$, $f^{\rm v}_{\rm enc}$, $f^{\rm au}_{\rm enc}$) take audio and visual signals (i.e., ${\bm X}_{\rm a}$ and ${\bm X}_{\rm v}$) as inputs and output corresponding latent representations, respectively. In particular, we introduce a facial action unit (FAU) encoder $f^{\rm au}_{\rm enc}$ designed to extract fine-grained structured geometric information from facial motions in video streams. Then, we apply a fusion operation $f_{\rm \varphi}$ to derive FAU-enhanced visual representations by integrating video features with FAU features. To better learn information interaction between modalities, we first perform implicit feature alignment across the temporal domain via our designed Query-shared audio/visual transformers $f_{\rm at}^{\rm q}$/$f_{\rm vt}^{\rm q}$. Then, we propose a temporal attentional pooler $f_{\rm \phi}$ in a frame-by-frame manner to capture temporal consistency via inter- and intra-modal modeling. Finally, we feed flattened pooled vectors into MLP-based classifiers $f_{\rm mlp}$ for audio-visual forgery predictions $\bm s$. The process described above can be formally expressed as follows:

    \begin{gather}
        {\bm s} = f_{\rm mlp}(f_{\rm \phi}(f_{\rm at}^{\rm q}(f_{\rm enc}^{\rm a}({\bm X}_{\rm a})),f_{\rm vt}^{\rm q}(f_{\rm \varphi}(f_{\rm enc}^{\rm v}({\bm X}_{\rm v}),f_{\rm enc}^{\rm au}({\bm X}_{\rm v}))))).
    \end{gather}

\subsection{FAU-enhanced Feature Learning}\label{subsection:ffl}

    For the visual modality, we first extract $T$ consecutive video frames ${\bm X}_{\rm v} \in \mathbb{R}^{T \times C \times H \times W}$  from the given video. To better enable the model to focus on fine-grained facial motion across the temporal domain, we utilize the pre-trained FAU encoder $f^{\rm au}_{\rm enc}$ with frozen weights for FAU feature extraction. FAU features can help forgery detectors capture FAU-related forgeries, thereby enhancing visual representation learning for audio-visual deepfake detection. The FAU latent features are obtained as ${\bm Z}_{\rm au}=f_{\rm enc}^{\rm au}({\bm X}_{\rm v}) $. Moreover, to learn richer visual representations, we use the video encoder $f^{\rm v}_{\rm enc}$ with learnable weights for further extracting video features ${\bm Z}_{\rm vid}=f_{\rm enc}^{\rm v}({\bm X}_{\rm v}) $. Finally, we utilize a feature fusion operation to obtain FAU-enhanced visual latent features ${\bm Z}_{\rm v}= f_{\rm \varphi}({\bm Z}_{\rm vid}, {\bm Z}_{\rm au}) $, which are dimensionally aligned with the audio latent features ${\bm Z}_{\rm a}\in \mathbb{R}^{T \times L}$.

    To better analyze audio signals, we extract a mel-spectrogram from raw audio waveforms. The resulting Mel-spectrogram $\bm X_{\rm a}$ is a 2-dimensional representation of sound signals. Therefore, Mel-spectrogram is essentially a single-channel image including time and frequency information simultaneously, which can be taken as the input of the audio encoder $f^{\rm a}_{\rm enc}$. The resulting audio latent features ${\bm Z}_{\rm a}$ can be denoted as ${\bm Z}_{\rm a}=f_{\rm enc}^{\rm a}({\bm X}_{\rm a}) $.

\subsection{Implicit Feature Alignment}\label{subsection:ifa}

    Before modeling the temporal correlation between audio and visual modalities, we propose a novel implicit feature alignment approach to learn the common information between the two modalities and reduce information redundancy, thereby enhancing the generalization of audio-visual deepfake detection. Motivated by~\cite{vaswani2017attention,li2023blip}, we design a Query-shared multimodal transformer $f_{\rm t}^{\rm q}(\bm Q,\bm K_{\rm a},\bm V_{\rm a}, \bm K_{\rm v}, \bm V_{\rm v}) = [{\bm Z}_{\rm aq}, {\bm Z}_{\rm vq}]$ with weight-shared learnable queries ${\bm Q} \in \mathbb{R}^{T \times L }$.

    Both audio and visual latent features, ${\bm Z}_{\rm a}$ and ${\bm Z}_{\rm v}$, can be mapped into Key and Value (${\bm K\& \bm V}$) pairs using $1\times1$ convolutions. The Query-shared Transformer (QT) operation is defined as follows:
    \begin{gather}
        {\bm Z}_{\rm aq} = {\rm softmax}({\bm Q}{\bm K_{\rm a}}^{\rm T}/\sqrt{d_{\rm a}}){\bm V_{\rm a}},  \\
        {\bm Z}_{\rm vq} = {\rm softmax}({\bm Q}{\bm K_{\rm v}}^{\rm T}/\sqrt{d_{\rm v}}){\bm V_{\rm v}},
    \end{gather}
    where $\sqrt{d_{\rm v}}$ and $\sqrt{d_{\rm a}}$ serve as scaling factors. The aligned features ${\bm Z}_{\rm aq} \in \mathbb{R}^{T \times L }$ and ${\bm Z}_{\rm vq} \in \mathbb{R}^{T \times L }$ facilitate the learning of inter-modal feature interactions.

\subsection{Temporal Attentional Pooler}\label{subsection:tap}
    To capture fine-grained interaction information across both intra- and inter-modalities, we attempt to model temporal consistency by calculating frame-wise correlations within intra- and inter-modalities. Temporal consistency is expected to generalize effectively to diverse and unseen forgery methods across different modalities. To this end, we design a Temporal Attentional Pooler that can produce a dense attentional matrix to measure the temporal consistency of intra- or inter-modalities. The dense attentional matrix ${\bm M}_{\rm av}\in \mathbb{R}^{T \times T }$ can be obtained from ${\bm Z}_{\rm aq} $ and ${\bm Z}_{\rm vq} $ as follows:
    \begin{gather}
        {\bm M}_{\rm av} = f_{\rm norm}(\sigma_{\rm av} * f_{\rm mp}({\bm Z}_{\rm aq}, {\bm Z}_{\rm vq})),
    \end{gather}
    where $\sigma_{\rm av} $ denotes a scaling factor with learnable parameters. $f_{\rm mp}$ and $f_{\rm norm}$ are the matrix product of two input tensors and the normalization operation, respectively.

    The above audio-visual dense attentional matrix ${\bm M}_{\rm av}$ can describe discriminative interaction information between the two modalities, which is expected to reveal inter-modal forgery traces. Similarly, we can obtain the audio/visual dense attentional matrix, ${\bm M}_{\rm a}$/${\bm M}_{\rm v}$, by modeling the intra-modal temporal relationship as follows:
    \begin{gather}
        {\bm M}_{\rm a} = f_{\rm norm}(\sigma_{\rm a} * f_{\rm mp}({\bm Z}_{\rm aq}, {\bm Z}_{\rm aq})), \\
        {\bm M}_{\rm v} = f_{\rm norm}(\sigma_{\rm v} * f_{\rm mp}({\bm Z}_{\rm vq}, {\bm Z}_{\rm vq})),
    \end{gather}
    where $\sigma_{\rm a}$ and $\sigma_{\rm v}$ denote scaling factors with learnable parameters for the audio and visual modalities, respectively.

\subsection{Multiple Forgery Detection}\label{subsection:mfd}

    Next, the dense attentional matrix of the inter-modality is flattened and fed into an independent multi-layer perceptron (MLP) $f^{\rm av}_{\rm mlp}$ for multimodal forgery prediction. The inter-modality prediction ${\bm s_{\rm av}}$ for the audio-visual deepfake detection can be formulated as:
    \begin{gather}
        {\bm s_{\rm av}} = f^{\rm av}_{\rm mlp}( f_{\rm \rho}({\bm M}_{\rm av}) ),
    \end{gather}
    where $f_{\rm \rho}$ indicates the flattened operation. Similarly, we can obtain visual-only and audio-only forgery predictions (i.e., ${\bm s_{\rm a}}$ and ${\bm s_{\rm v}}$) for the unimodal deepfake detection. The predicted scores for the audio and visual modalities are obtained as follows:
    \begin{gather}
        {\bm s_{\rm a}} = f^{\rm a}_{\rm mlp}( f_{\rm \rho}({\bm M}_{\rm a}) ), \\
        {\bm s_{\rm v}} = f^{\rm v}_{\rm mlp}( f_{\rm \rho}({\bm M}_{\rm v}) ),
    \end{gather}    
    where $f^{\rm a}_{\rm mlp}$ and $f^{\rm v}_{\rm mlp}$ denote the audio-only classifier and visual-only classifier, respectively.

    Finally, we employ the standard cross-entropy loss function $f_{\rm ce}(\cdot, \cdot)$ to compute the loss $\mathcal{L}_{\rm *}=f_{\rm ce}({\bm y}, {\bm s})$ between the ground-truth labels ${\bm y}$ and predicted scores ${\bm s}$. The total loss function can be obtained as follows:
    \begin{gather}
        \mathcal{L}_{\rm total} = \lambda_{\rm av} \mathcal{L}_{\rm av} +  \lambda_{\rm a} \mathcal{L}_{\rm a} +\lambda_{\rm v} \mathcal{L}_{\rm v},
    \end{gather}
    where $\lambda_{\rm av}$, $\lambda_{\rm a}$, and $\lambda_{\rm v}$ denote the weight parameters used to balance the contributions of each loss.
    
    During the inference phase, the multimodal prediction score ${\bm s_{\rm av}}$ is used as the final output.

\subsection{Implementation Details}\label{subsection:id}
    Specifically, our video encoder, FAU encoder, and audio encoder are derived from the backbone network of CSN~\cite{tran2019video}, ME-GraphAU~\cite{LuoS0SG22}, and Whisper~\cite{radford2023robust}, respectively. The FAU encoder is pre-trained on the DISFA dataset~\cite{mavadati2013disfa}, which provides fine-grained FAU annotations essential for learning subtle facial motion patterns. The MLP module is only a single-layer fully-connected layer with 512 neurons. In our framework, each input video sample consists of $T=25$ frames. The dimension of the learnable queries ${\bm Q} \in \mathbb{R}^{T \times L }$ used in our Query-shared Transformer is $25\times 512$. The trade-off parameters $\lambda_{\rm a}$, $\lambda_{\rm v}$, and $\lambda_{\rm av}$, which balance the contributions of the audio, visual, and audio-visual losses, are empirically set to $0.1$, $0.1$, and $0.8$, respectively.

\section{Experiments}\label{section:exp}
    In this section, we evaluate the performance of our proposed method on two publicly available audio-visual deepfake datasets (i.e., LAV-DF~\cite{cai2022you} and FakeAVCeleb~\cite{khalid2021fakeavceleb}) for within- and cross-database evaluations.
    
\subsection{Datasets and Settings}\label{subsection:settings}

    \noindent\textbf{Dataset.} FakeAVCeleb (FakeAV for short) is a widely used multimodal forgery dataset for audio-visual deepfake detection. It is constructed using a combination of visual manipulation techniques and voice cloning methods, based on 500 English-speaking video samples. The dataset comprises 500 real videos and over 20,000 forged ones. Due to the absence of an official train/test split, we adopt an identity-independent train/test splitting strategy, following the protocol described in~\cite{zou2024cross}.

    LAV-DF is a newly introduced large-scale audio-visual deepfake dataset established for temporal forgery detection and localization. It contains 36,431 real videos and 99,873 fake videos, and provides an official train/test split for comprehensive performance evaluation of detectors.

\begin{figure}[t]
  \centering
   \includegraphics[width=1\linewidth]{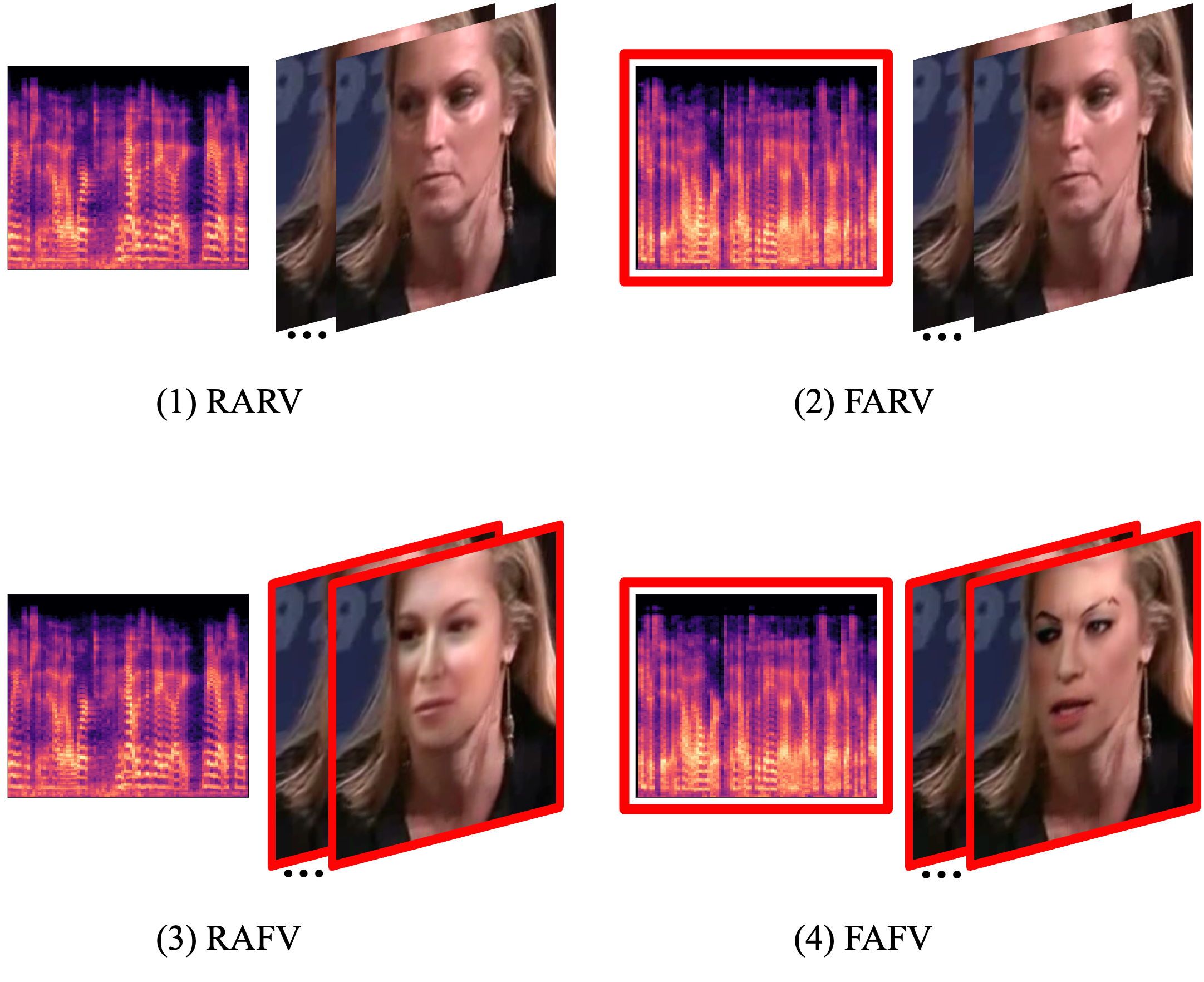}
   \caption{Samples from four categories of audio-visual signals. Forged samples are highlighted with a \red{red box}.}
   \label{fig:database}
\end{figure}

    Both the FakeAV and LAV-DF datasets include four types of audio-visual combinations: RealAudio-RealVideo ($RARV$), FakeAudio-RealVideo ($FARV$), RealAudio-FakeVideo ($RAFV$), and FakeAudio-FakeVideo ($FAFV$). Representative samples from these four categories are illustrated in Fig.~\ref{fig:database}. In addition, we leverage the well-known DISFA dataset~\cite{mavadati2013disfa} to pre-train our FAU encoder. DISFA consists of 130,788 facial images annotated with action unit (AU) labels and is widely adopted for facial action unit recognition tasks.

    \noindent\textbf{Data Preparation.} To focus on mouth movements, we apply a center crop to extract a $160\times 160$ facial region for training our proposed model. For audio processing, we compute log-Mel spectrograms using a 25-ms Hann window with a hop of 10 ms and 80 Mel filter banks as acoustic input features. The resulting mel-spectrogram is a 2D array with the size of $80\times 100$. We take 25 consecutive video frames along with the corresponding audio signals as the input of the model. In addition, we employ general data augmentation methods, following RealForensics~\cite{hal2022l}.

    \begin{table*}[h]
      \centering
      \caption{  Binary classification performance comparisons among 12 different forgery detectors on FakeAV and LAV-DF under within-database evaluations.}  %
        \begin{tabular}{ccccccccccc}
        \bottomrule
    
        \multirow{2}{*}{}  &\multirow{2}{*}{}  &\multirow{2}{*}{}  &\multicolumn{2}{c}{FakeAV} &\multicolumn{2}{c}{LAV-DF}  &\multicolumn{2}{c}{AVG}\\
          \cmidrule(r){4-5} \cmidrule(r){6-7} \cmidrule(r){8-9} \cmidrule(r){10-11} 
          Model  & Modality    & Category   & \multicolumn{1}{c}{Acc} & \multicolumn{1}{c}{AUC} & \multicolumn{1}{c}{Acc} & \multicolumn{1}{c}{AUC}  & \multicolumn{1}{c}{Acc} & \multicolumn{1}{c}{AUC} \\
        \bottomrule
    LipForensics~\cite{h2021lips}                            & V  & Unimodal   & 91.96 & 96.79 & 73.18 & 79.74 & 82.57  & 88.27  \\
    RealForensics~\cite{hal2022l}                            & V  & Unimodal   & 97.33 & 98.78 & 74.83 & 71.77 & 86.08  & 85.28  \\
    RawGAT~\cite{tak2021end}  & A  & Unimodal   & 55.09 & 77.97 & 73.96 & 82.52 & 64.53  & 80.25  \\
    SSLAS~\cite{tak2022automatic}  & A  & Unimodal   & 51.31  & 76.24 & 72.30 & 94.31 & 63.78  & 85.28  \\
    
    LipForensics~\cite{h2021lips}$+$RawGAT~\cite{tak2021end} & AV & Ensemble   & 56.21 & 98.49 & 70.95 & 99.85 & 63.58  & 99.17\\
    RealForensics~\cite{hal2022l}$+$RawGAT~\cite{tak2021end} & AV & Ensemble   & 57.77 & 99.91 & 64.03 & 99.88 & 60.90  & 99.90 \\
    EnsembleAVDF~\cite{hashmi2022multimodal}                 & AV & Ensemble   & 98.29 & 98.72 & 74.17 & 99.90 & 86.23  & 99.31   \\
    AVD-DDL~\cite{chugh2020not}                              & AV & Multimodal & 97.81 & 99.66 & 99.28  & 99.92 & 98.55  & 99.79   \\  
    AVTS~\cite{sung2023hearing}                              & AV & Multimodal & 97.30 & 96.89 & 69.13  & 77.61 & 83.22  & 87.25   \\
    MRDF~\cite{zou2024cross}                                 & AV & Multimodal & 94.77 & 90.36 & 98.47  & 99.80 & 96.62  & 95.08   \\
    FGMDF~\cite{yin2024fine}                                 & AV & Multimodal & 99.66 & 99.84 & \textbf{99.79}  & 99.96 & 99.73  & 99.90   \\
    
        \hline
        FauForensics (Ours)           & AV & Multimodal   & \textbf{99.71} & \textbf{99.91}   & 99.78  & \textbf{99.97}  &    \textbf{99.75}  & \textbf{99.94} \\

        \bottomrule
        \end{tabular}
      \label{table_within2}
    \end{table*}%

    \begin{table*}[htbp]
      \centering
      \caption{ Four-class classification performance comparison among 4 different forgery detectors on FakeAV and LAV-DF under within-database evaluations.}  %
        \begin{tabular}{ccccccccccc}
        \bottomrule
    
        \multirow{2}{*}{}  &\multirow{2}{*}{}  &\multirow{2}{*}{}  &\multicolumn{2}{c}{FakeAV} &\multicolumn{2}{c}{LAV-DF}  &\multicolumn{2}{c}{AVG}\\
          \cmidrule(r){4-5} \cmidrule(r){6-7} \cmidrule(r){8-9} \cmidrule(r){10-11} 
          Model  &  Modality  & Category   & \multicolumn{1}{c}{Acc} & \multicolumn{1}{c}{AUC} & \multicolumn{1}{c}{Acc} & \multicolumn{1}{c}{AUC}  & \multicolumn{1}{c}{Acc} & \multicolumn{1}{c}{AUC} \\
        \bottomrule
        AVTS~\cite{sung2023hearing}      & AV & Multimodal   & 96.56  & 99.58  & 67.70  & 87.17 & 82.13  & 93.38   \\
        MRDF~\cite{zou2024cross}         & AV & Multimodal   & 74.96  & 90.14  & 96.69  & 99.76 & 85.83  & 94.95   \\
        FGMDF~\cite{yin2024fine}         & AV & Multimodal   & 99.45  & 99.96  & 74.18  & 91.14 & 86.82  & 95.55   \\
        \hline
        FauForensics (Ours)                 & AV & Multimodal   & \textbf{99.47} & \textbf{99.97}  & \textbf{99.69}   &  \textbf{99.95}  &    \textbf{99.58}  & \textbf{99.96} \\
        \bottomrule
        \end{tabular}
      \label{table_within4}
    \end{table*}%

    \noindent\textbf{Compared Methods.} To demonstrate the effectiveness of the proposed method, we compare our method with 9 advanced deepfake detectors including 4 unimodal detectors (i.e., LipForensics~\cite{h2021lips}, RealForensics~\cite{hal2022l}, RawGAT~\cite{tak2021end}, SSLAS~\cite{tak2022automatic}), 3 ensemble detectors (i.e., LipForensics~\cite{h2021lips}$+$RawGAT~\cite{tak2021end}, RealForensics~\cite{hal2022l}$+$RawGAT~\cite{tak2021end}, and EnsembleAVDF~\cite{hashmi2022multimodal}), and 4 multimodal detectors (i.e., AVD-DDL~\cite{chugh2020not}, AVTS~\cite{sung2023hearing}, MRDF~\cite{zou2024cross}, and FGMDF~\cite{yin2024fine}). It should be noted that AVD-DDL~\cite{chugh2020not}, a multimodal deepfake detector, predicts authenticity based on a modality dissonance score threshold, making it suitable only for binary classification tasks and unsuitable for four-class classification. Similarly, the unimodal and ensemble detectors are also limited to binary classification and can not be directly applied to four-class classification scenarios.

    \noindent\textbf{Training Details.} We implement our proposed method on the PyTorch platform and utilize the AdamP~\cite{heo2020adamp} optimizer to update the network weights. The initial learning rate, mini-batch size, and number of training epochs are set to 1e-4, 32, and 50, respectively. Besides, we employ a poly learning rate policy with a power of 0.9 to adjust the learning rate during training.

    \noindent\textbf{Evaluation Metrics.} In this work, we adopt two widely used evaluation metrics in the deepfake detection field, i.e., Accuracy (Acc for short) and Area Under the ROC Curve (AUC). Since our proposed method operates at the video level, we report experimental results for all video-level detectors based on the video clip-level ground truth. For ensemble detectors, we obtain multimodal predictions by averaging the outputs of the corresponding unimodal detectors.

\subsection{Within-database Evaluation}\label{subsection:within}

    Most existing work usually treats unimodal deepfake detection as a binary classification task, where the objective is to distinguish between authentic and forged content. However, since audio and visual modalities can be manipulated either independently or jointly in multimodal deepfakes, simple binary classification is insufficient. This is because binary classification fails to provide insights into the specific category of multimodal forgery. As a result, it restricts the detector's interpretability and reduces its usefulness in forensic analysis, where understanding the exact type of manipulation is crucial for informed decision-making and further investigation. To comprehensively analyze the types of forgery in multimodal deepfakes, we conduct both binary and four-class classification in within-database and cross-database evaluation. The four-class classification setting includes the categories: RealVideo–RealAudio, RealVideo–FakeAudio, FakeVideo–RealAudio, and FakeVideo–FakeAudio. Table~\ref{table_within2} and Table~\ref{table_within4} report the binary and four-class classification results on the FakeAV and LAV-DF datasets, respectively. In both tables, the last column labeled ``AVG'' represents the average performance across FakeAV and LAV-DF evaluations (the same as below unless indicated).

\begin{figure}[H]
    \centering
    \subfloat[AVTS]{%
        \includegraphics[width=0.5\linewidth]{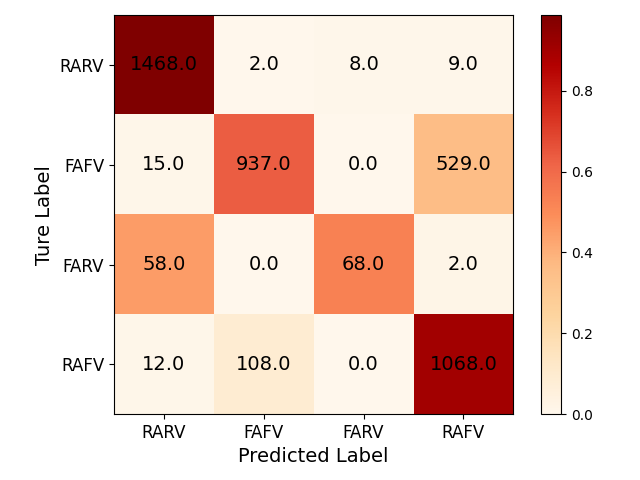}%
        \label{subfig:AVTS}%
    }\hfill
    \subfloat[MRDF]{%
        \includegraphics[width=0.5\linewidth]{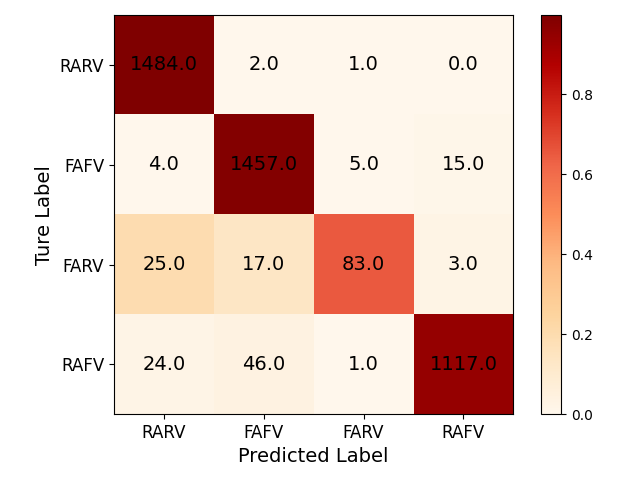}%
        \label{subfig:MRDF}%
    }
    \hfill
    \subfloat[FGMDF]{%
        \includegraphics[width=0.5\linewidth]{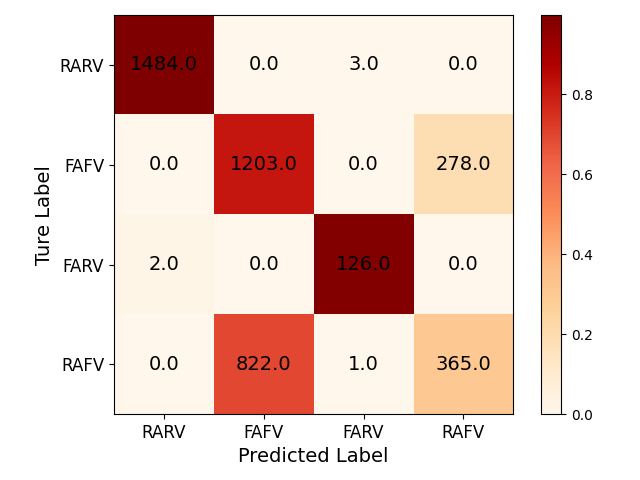}%
        \label{subfig:FGMDF}%
    }
    \hfill
    \subfloat[Ours]{%
        \includegraphics[width=0.5\linewidth]{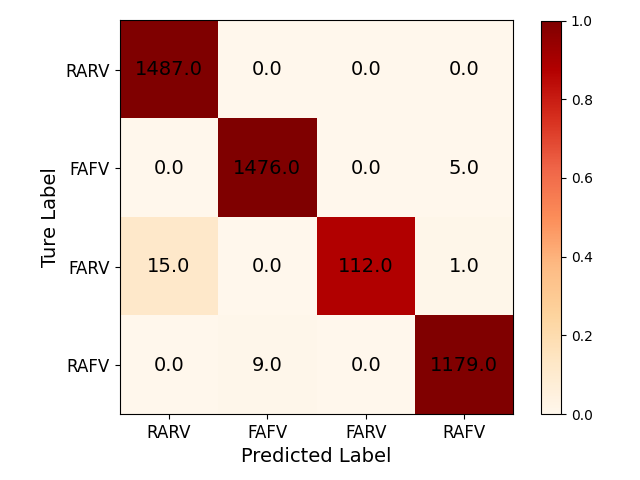}%
        \label{subfig:Ours}%
    }    
    \caption{The confusion matrix of four multimodal forgery detectors in four-class classification evaluations. Each element in the matrix represents the number of samples predicted for the corresponding class. For a fair comparison, all four detectors are evaluated using the same test list employed by FGMDF. } 
    \label{cm} 
\end{figure}

    As shown in Table~\ref{table_within2}, we compare our proposed method with four unimodal detectors, three ensemble-based detectors, and four multimodal detectors in within-database evaluations. We can see that our proposed method outperforms all other detectors in terms of average performance across the FakeAV and LAV-DF datasets. Unimodal detectors struggle with multimodal forgeries due to the absence of multimodal information, making it difficult to capture all the artifacts from both audio and visual streams. Although ensemble-based detectors aggregate predictions from unimodal models, they lack explicit modeling of inter-modal interactions, which hampers their ability to detect complex cross-modal manipulations effectively. In contrast, multimodal detectors are more capable of capturing cross-modal inconsistencies, leading to stronger performance in binary within-database evaluations. Notably, both our method and FGMDF achieve top average results for binary classification across both datasets. However, it is worth noting that FGMDF processes 4-second video clips, which leads to the exclusion of shorter videos during evaluation. A fairer comparison between our approach and FGMDF, accounting for these differences, is presented in Fig.~\ref{cm}.

    Table~\ref{table_within4} presents performance comparisons among 4 different forgery detectors for the four-class classification task on the FakeAV and LAV-DF datasets. Our proposed method also outperforms all the deepfake detectors across all evaluation metrics in four-class classification scenarios. Although FGMDF achieves competitive results in binary classification scenarios, it suffers from obvious performance degradation for four-class classification, particularly on the LAV-DF dataset. In contrast, our method maintains robust performance and achieves the best overall results in multiple classification scenarios. Specifically, compared to the current state-of-the-art (i.e., FGMDF), our approach yields an average improvement of 4.41\% in AUC and 12.76\% in accuracy. These results demonstrate the effectiveness of our method in accurately detecting various types of forgeries under within-database evaluation settings.

  \begin{table*}[ht]
    \centering
    \caption{ Performance comparison among 5 different multimodal detectors for the binary and four-class classification using the AUC metric in cross-database evaluations. Note that ``FakeAV" in the column refers to the setting where the model is trained on LAV-DF and tested on FakeAV, and vice versa.}  %
      \begin{tabular}{ccccccccc}
      \bottomrule
  
      \multirow{2}{*}{}  &\multirow{2}{*}{}  &\multirow{2}{*}{}  &\multicolumn{3}{c}{Binary} &\multicolumn{3}{c}{Four-class}\\
        \cmidrule(r){4-6} \cmidrule(r){7-9}  
        Model  &  Modality  &  Category  & \multicolumn{1}{c}{FakeAV} & \multicolumn{1}{c}{LAV-DF} & \multicolumn{1}{c}{Avg} & \multicolumn{1}{c}{FakeAV} & \multicolumn{1}{c}{LAV-DF} & \multicolumn{1}{c}{Avg}   \\
      \bottomrule
      AVD-DDL~\cite{chugh2020not}      & AV & Multimodal & 66.03  & 77.44  & 71.74  & - & -  & -   \\  
      AVTS~\cite{sung2023hearing}      & AV & Multimodal & 60.35  & 64.90  & 62.63  & 67.72 & 73.63  & 70.68   \\
      MRDF~\cite{zou2024cross}         & AV & Multimodal & 81.53  & 67.47  & 74.50  & 67.20 & 66.69  & 66.95   \\
      FGMDF~\cite{yin2024fine}         & AV & Multimodal & 78.27  & 91.28  & 84.78  & 61.80 & 65.10  & 63.45   \\
      
      \hline
      FauForensics (Ours)           & AV & Multimodal   & \textbf{83.77} & \textbf{95.44} &  \textbf{89.61} & \textbf{71.20}  & \textbf{92.10}   & \textbf{81.65} \\

      \bottomrule
      \end{tabular}
    \label{table_cross}
  \end{table*}%

    To gain deeper insights into the experimental results, we visualize the confusion matrix of four-class classification results on the test set under the within-database evaluation setting, as shown in Fig.~\ref{cm}. To perform a fair comparison between multiple multimodal detectors, we use the same test list as FGMDF on the LAV-DF dataset. As illustrated in Fig.~\ref{cm}, we can see that FGMDF easily misclassifies $RAFV$ samples as $FAFV$, indicating a limitation in distinguishing between visual-only forgeries and fully forged audio-visual content. In contrast, our method achieves a much higher accuracy in identifying $RAFV$ instances. These observations further highlight the superiority of our approach over existing state-of-the-art detectors in capturing complex multimodal forgery patterns.

  \subsection{Cross-database Evaluation}\label{subsection:cross}
    Cross-database evaluation has been widely regarded as one of the most challenging tasks in the field of deepfake detection. Unlike within-database evaluation, which assesses the performance of detectors on the same dataset used for both training and testing, cross-database evaluation involves training on one dataset and evaluating on another unseen dataset. This setting is recognized as a reliable benchmark for evaluating the generalization ability of deepfake detectors, especially their robustness to previously unseen manipulation techniques and post-processing operations. Generally, most deepfake detectors can achieve promising results under within-database conditions but tend to exhibit substantial performance degradation when applied to new datasets. This performance drop indicates that deepfake detectors are likely overfitting to dataset-specific artifacts rather than learning generalizable forgery-relevant features. In this section, we provide a detailed comparison of different audio-visual deepfake detectors under cross-database evaluation, as summarized in Table~\ref{table_cross}. Our analysis focuses on the models’ ability to maintain high detection performance when faced with real-world variability, including diverse forgery patterns and different dataset domains.

    Table~\ref{table_cross} presents the performance (AUC) comparison of the cross-database evaluation under both binary and four-class classification settings. The experiments include two cross-database scenarios: 1) training on LAV-DF and testing on FakeAV, and 2) training on FakeAV and testing on LAV-DF. As shown in the table~\ref{table_cross}, our proposed method achieves state-of-the-art performance across both classification settings. Specifically, in the binary classification scenario, our method outperforms the previous SOTA audio-visual deepfake detector (i.e., FGMDF) by an average AUC improvement of 5.74\% (90.52\% vs. 84.78\%). In the more challenging four-class classification scenario, all detectors suffer from an obvious performance drop due to the increased task complexity, similar to observations in within-database evaluations. However, our method exhibits a smaller performance degradation compared to other detectors and still achieves the highest average AUC across evaluations. These results demonstrate that our FAUs-guided framework not only captures subtle multimodal inconsistencies effectively but also enhances cross-dataset generalizability. This robustness against unseen forgeries and dataset domain further highlights the generalizability and reliability of our approach in real-world applications.

  \begin{figure*}[htbp]
    \centering
    \includegraphics[width=1.0\linewidth]{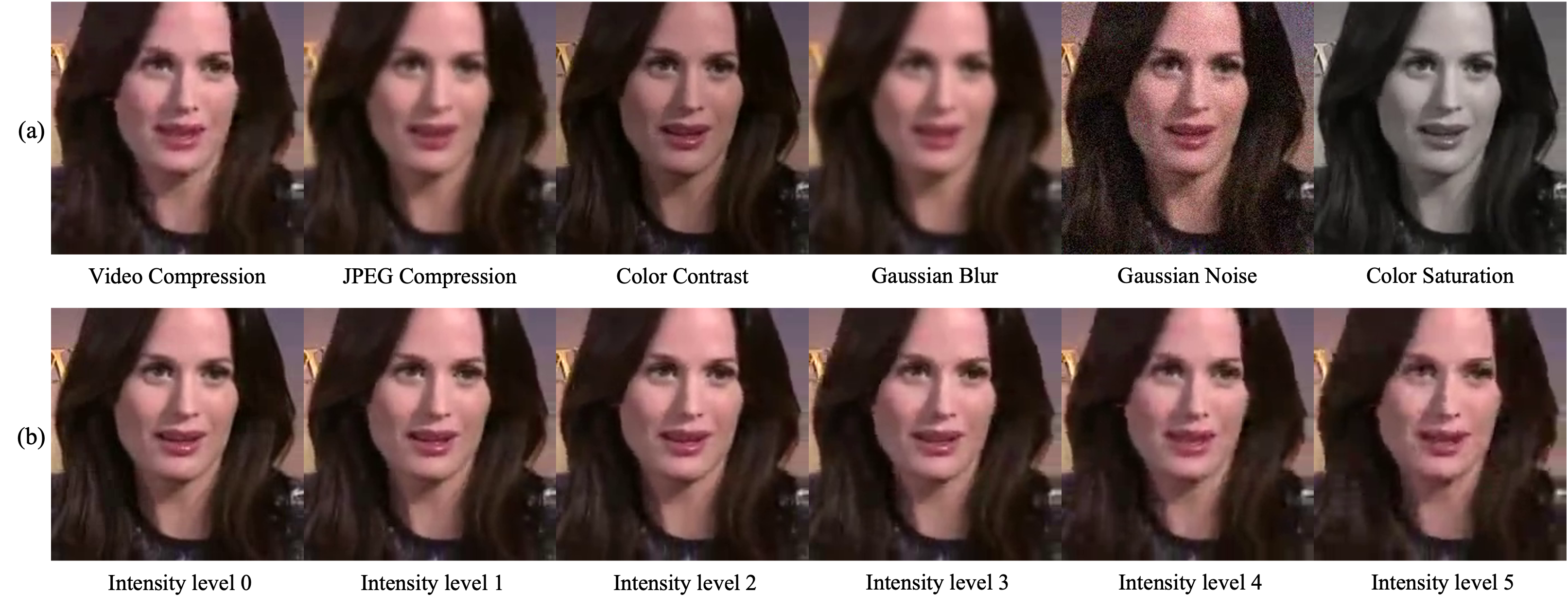}
    \caption{ (a) Examples of facial images under different perturbations. (b)  Examples of video compression perturbations at different intensity levels. Higher intensity levels indicate stronger perturbations.}
  \label{p_example}
  \end{figure*}

  \begin{figure*}[htbp]
    \centering
    \includegraphics[width=1.0\linewidth]{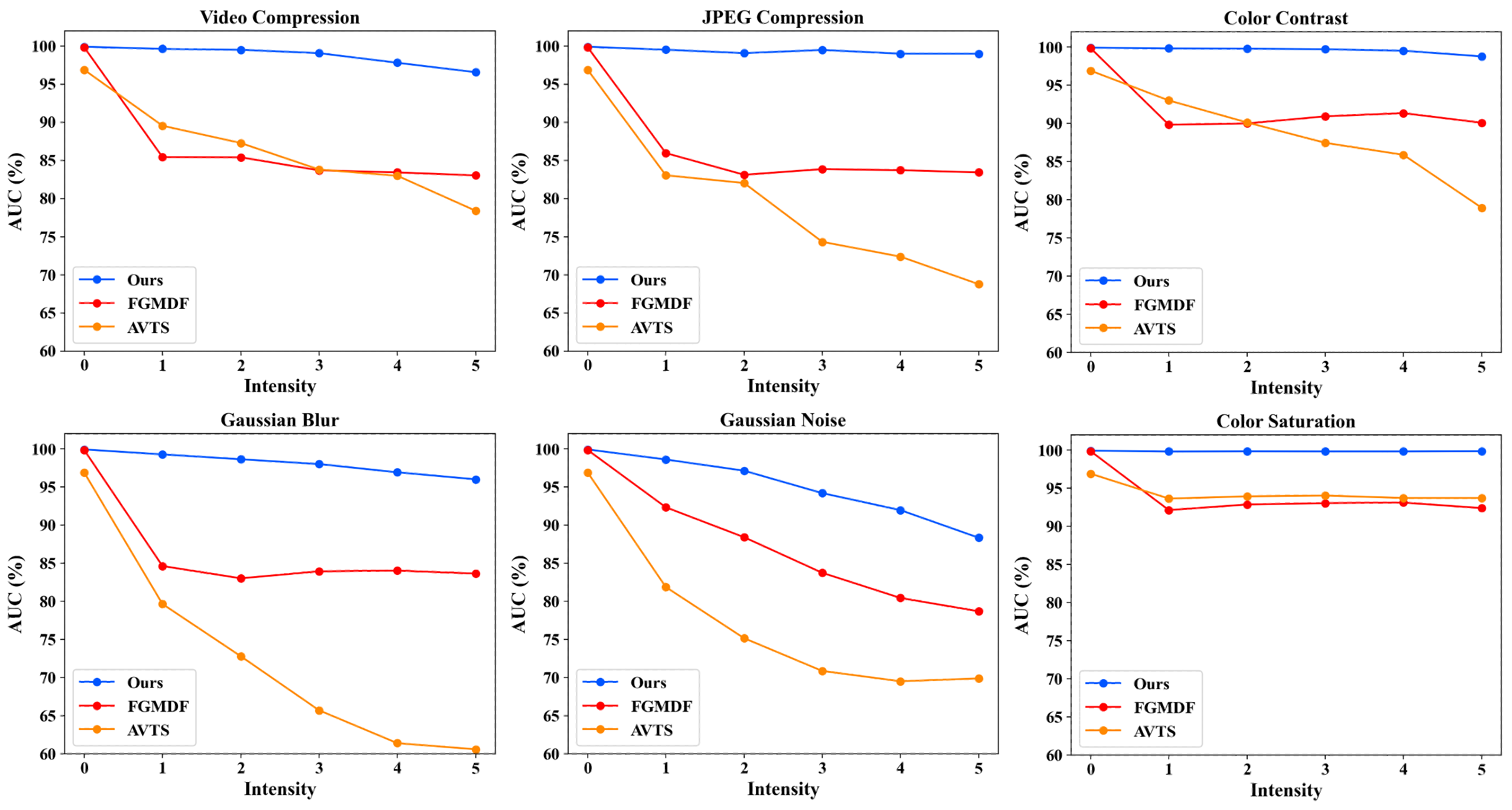}
    \caption{ Robustness to unseen perturbation on FakeAV. AUC scores (\%) are reported across five intensity levels for six types of perturbations: video compression, JPEG compression, color contrast, Gaussian blur, Gaussian noise, and color saturation. }
  \label{robust}
  \end{figure*}

  \subsection{Robustness to Perturbations}\label{subsection:pert}

    A common challenge arises from widespread post-processing perturbations, such as image/video compression, noise, etc. These perturbation operations can degrade manipulation artifacts, thereby compromising the robustness and reliability of forgery detection systems. In this study, we consider the six perturbation methods, including Video Compression, JPEG Compression, Color Contrast, Gaussian Blur, Gaussian Noise, and Color Saturation, as presented in Fig.~\ref{p_example}(a). Following ~\cite{jiang2020deeperforensics}, we systematically apply each perturbation to the original video at five different intensity levels. Fig.~\ref{p_example}(b) presents examples of video compression at different intensity levels. Note that Intensity level 0 corresponds to the original video without any perturbations. To evaluate robustness against perturbations, we conduct a comprehensive experimental evaluation comparing our approach with two prior SOTA methods across various types of perturbations, as shown in Fig.~\ref{robust}. The experimental results clearly demonstrate that our method achieves excellent robustness against all perturbation types and intensity levels. The superior robustness of our method stems from the FAU-enhanced cross-modal modeling mechanism, which facilitates effective feature learning even under severe post-processing distortions.

\subsection{Ablation Study}\label{subsection:as}

    In this section, we conduct two groups of ablation experiments to analyze the effect of the core modules and the feature extractors. 

    \textbf{Core Modules. } To systematically evaluate the contribution of each core module in our framework, we conduct an ablation study by comparing three variants with the full proposed method: Variant-A, Variant-B, and Variant-C, as summarized in Table~\ref{table_as1}. Specifically, our framework consists of three key modules: (1) FAU-enhanced Feature Learning (FFL) module, (2) Implicit Feature Alignment (IFA) module, and (3) Temporal Attentional Pooler (TAP). The ablation experiments are performed by training on the LAV-DF dataset and testing on both FakeAV and LAV-DF, enabling evaluation under both within-database and cross-database scenarios. The quantitative results clearly demonstrate that each module contributes to overall performance improvement, validating the effectiveness of our designed module.

    \begin{table}[htbp]
    \centering
    \caption{Ablation study results of different variants of our framework in both within- and cross-database evaluations.  }  

      \begin{tabular}{cccccc}
        \bottomrule
        \multirow{2}{*}{}  &\multirow{2}{*}{}  &\multirow{2}{*}{} &\multirow{2}{*}{} &\multicolumn{2}{c}{Training on LAV-DF}  \\
        \cmidrule(r){5-6}   
        Model  &  FFL  & IFA   & TAP & \multicolumn{1}{c}{LAV-DF} & \multicolumn{1}{c}{FakeAV }   \\
        \bottomrule

        Variant-A   &             &             &             & 99.93   & 77.29   \\ 
        Variant-B   & \checkmark  &             &             & 99.95   & 79.33   \\ 
        Variant-C   & \checkmark  & \checkmark  &             & 99.95   & 80.97   \\ 
        \hline
        Ours        & \checkmark  & \checkmark  & \checkmark  & \textbf{99.97}     & \textbf{83.77}     \\ %
        
        \bottomrule
      \end{tabular}
    \label{table_as1}
  \end{table}%

    In Table~\ref{table_as1}, we can see that both our method and its variants obtain promising results in within-database evaluations. It should be noted that each module gradually improves the performance of cross-dataset evaluation. Specifically, Variant-B, which incorporates the FAU-enhanced Feature Learning (FFL) module, achieves about 2.04\% AUC (i.e., 79.33 vs. 77.29) improvement over Variant-A in cross-database evaluation. This can illustrate that the FFL module can enhance the discriminative feature learning and promote the performance improvement of audio-visual deepfake detection. In addition, benefit from the IFA module, Variant-C performs implicit feature alignment for better inter-modal learning and obtains 1.64\% AUC improvement over Variant-B. Finally, TAP centers on fine-grained frame-wise audiovisual correlations, further boosting detection performance and contributing to the overall robustness of our approach.

    \textbf{Feature Extractors.} Our proposed architecture contains three feature extractors: a FAU encoder, a Video encoder, and an Audio encoder. To evaluate the effectiveness of each feature extractor in our architecture, we compare our full model with three variants, as presented in Table~\ref{table_as2}. The quantitative results clearly demonstrate the contribution of each feature extractor to the overall performance. In particular, the variant without the FAU encoder suffers from significant performance degradation in cross-dataset evaluations, quantitatively confirming the essential role of facial action units in enhancing the generalization capability of our architecture.
  
  \begin{table}[htbp]
    \centering
    \caption{Ablation results regarding different encoders in both within- and cross-database evaluations. }  
    \begin{tabular}{lcc}
        \bottomrule
        \multirow{2}{*}{}  &\multicolumn{2}{c}{Training on LAV-DF}  \\
        \cmidrule(r){2-3}   
        \multicolumn{1}{c}{Model}  &  \multicolumn{1}{c}{LAV-DF} & \multicolumn{1}{c}{FakeAV}   \\
        \bottomrule

        w/o FAU encoder       &  99.96           & 78.01              \\ 
        w/o Video encoder     &  99.84           &  75.84             \\ 
        w/o Audio encoder     &  94.13           &  66.42             \\ 
        \hline
        FauForensics (Ours)   &  \textbf{99.97}  & \textbf{83.77}     \\ %
        
        \bottomrule
    \end{tabular}
    \label{table_as2}
  \end{table}%

  \begin{figure}[htbp]
    \centering
    \includegraphics[width=1\linewidth]{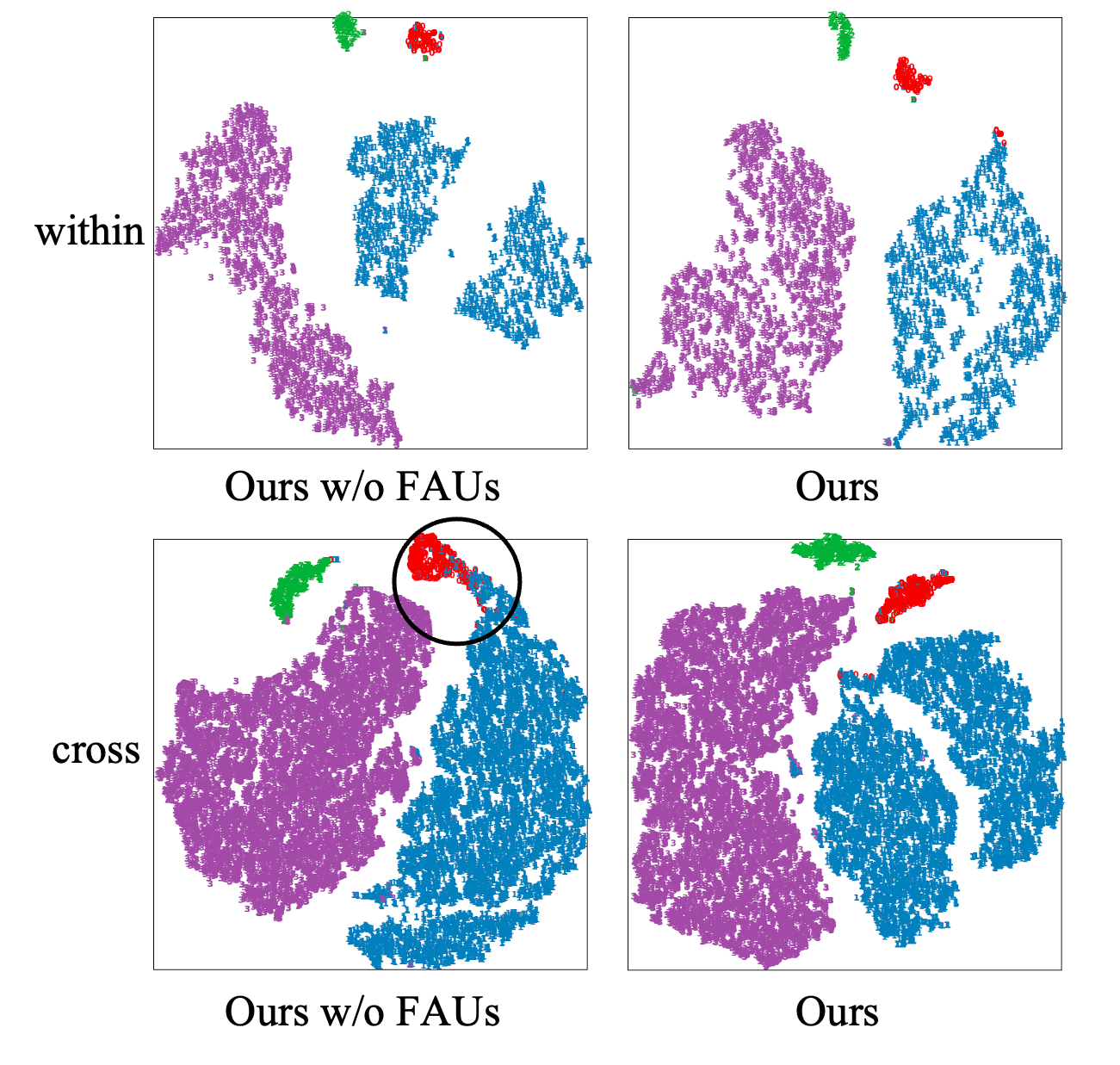}
    \caption{ The t-SNE visualization of features derived from our proposed method and ours w/o FAUs on the test sets of both within-database and cross-database evaluations. Red, blue, green, and violet indicate \textcolor{red}{$RARV$}, \textcolor{blue}{$RAFV$}, \textcolor{green}{$FARV$}, and \textcolor{violet}{$FAFV$}, respectively.}
  \label{fig:tsne}
  \end{figure}

  \begin{figure}[htbp]
    \centering
    \includegraphics[width=1\linewidth]{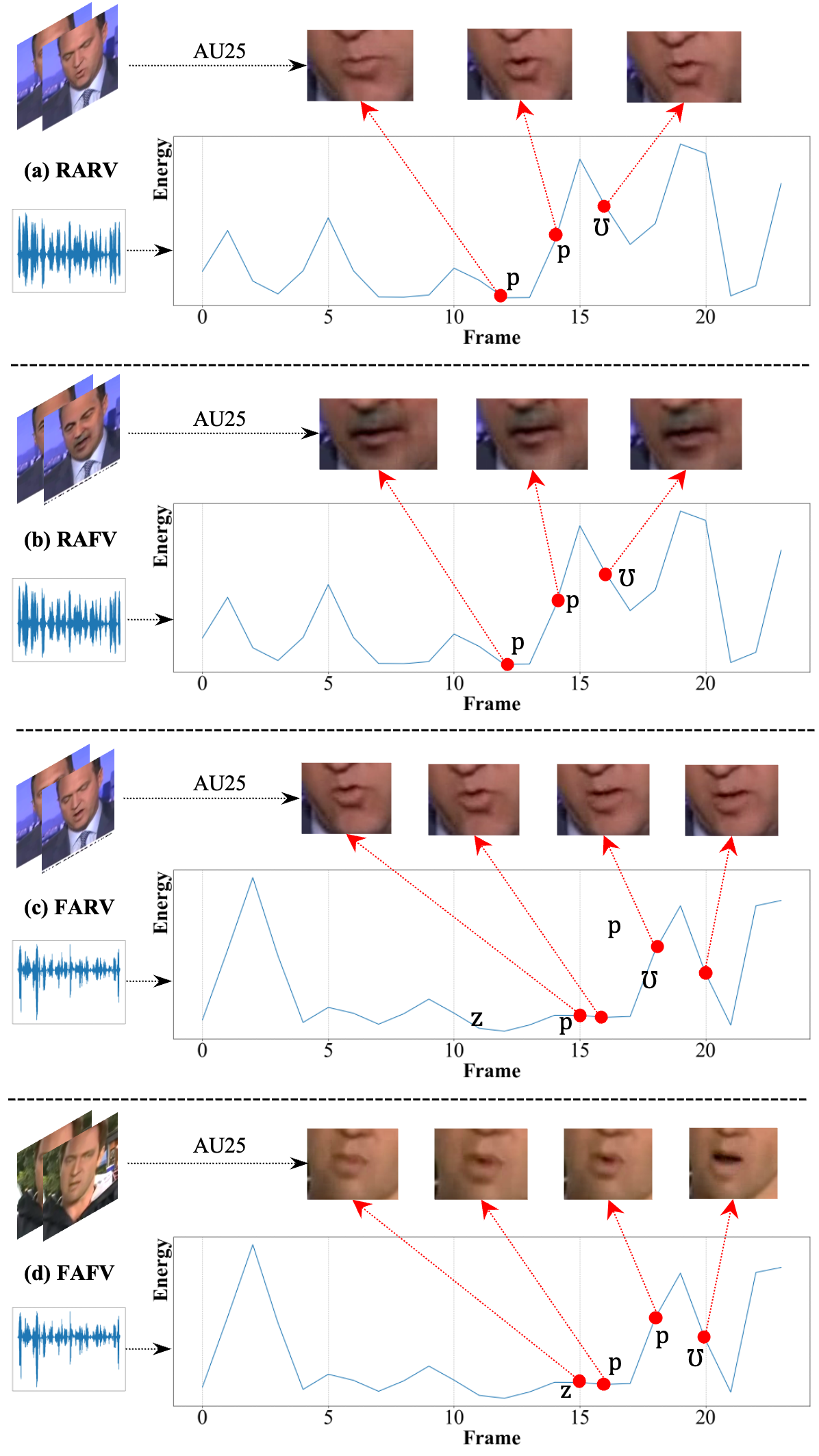}
    \caption{ Visualization of phoneme-to-AU25 correspondence. Four types of audio-visual pairs are presented: RARV, RAFV, FARV, and FAFV. Note that the phonemes $/\rm p/$  and $/\mho/$ occur in the word ``pulled''.  }
  \label{fig:pho}
  \end{figure}

\subsection{Visualization Analysis}\label{subsection:visual}

  To further illustrate the effectiveness of FAUs, we utilize t-distributed stochastic neighbor embedding (t-SNE)~\cite{van2008visualizing} to visualize features derived from our proposed method and ours w/o FAUs on the test set of within-database and cross-database (training on FakeAV), as depicted in Fig.~\ref{fig:tsne}. It can be observed that both methods obtain promising results in within-database scenarios, but the features of ours are more separable than those of ours w/o FAUs in cross-database scenarios. The circled part of Fig.~\ref{fig:tsne}(c) shows the weakened discriminability due to the lack of FAUs. Ours without FAUs can not distinguish $RARV$ (red part in Fig.~\ref{fig:tsne}) and $RAFV$ (blue part) well. However, our method benefits from the incorporation of FAUs, which can force the model to focus on more fine-grained facial motion cues, thereby enhancing its discriminative capability for audio-visual deepfake detection. Therefore, the above observation demonstrates that FAUs can boost feature learning of detectors.

  To better understand the role of FAU in the audio-visual forgery detection task, we visualize the phoneme-FAU pairings from four groups of audio-visual signals in Fig.~\ref{fig:pho}. 
  Facial action units describe nearly all possible facial actions. For example, AU25 (lip part) can depict the mouth motion, which is closely related to pronunciation. Typically, the pronunciation of certain phonemes is closely linked to the shape of the mouth. For instance, the pronunciation of the phoneme $/\rm p/$ requires the mouth to be tightly closed first. Therefore, we focus on the phoneme $/\rm p/$ and AU25 to show frame-level audio-visual forgery cues. In Fig.~\ref{fig:pho}, we observe that real audio-visual signals display the correct pronunciation of $/\rm p/$ in the word ``pulled'', which is to close the mouth first and then open it. However, the forged visual signals do not exhibit appropriate mouth movements corresponding to the phoneme at that moment for the RAFV pairing. Similarly, FARV did not follow this pronunciation mechanism. Both sets of visual signals indicate that the mouth remains slightly open, which does not match the pronunciation of $/\rm p/$. Finally, although FAFV improves lip-syncing via wavtolip, subtle traces imperceptible to the human eye can still be found in the frame-by-frame analysis of phoneme-FAU. For example, FAFV also exhibits a close-to-open mouth motion, with the mouth closure delayed by one audio frame.

\section{Conclusion}\label{section:conc}
    
    In this work, we propose a novel FAU-guided framework with frame-level integration to simultaneously capture various forgeries (i.e., audio, visual, and multi-modal forgeries). More precisely, we first introduce FAUs to learn facial geometric information for capturing FAU-related forgeries.  Moreover, a frame-level integration module is designed to capture intra- and inter-modal temporal correlations from audio-visual signals. Extensive experiments on two widely used audio-visual deepfake databases illustrate the superiority of our proposed method to prior SOTA methods.

    \noindent\textbf{Limitations.} Despite the promising performance, our method has some limitations. For example, our detection framework relies on the availability of both audio and visual streams. This requirement also restricts the method's applicability to scenarios involving incomplete or single-modal media, such as GIF and muted videos.  Moreover, our approach requires that the audio is clearly aligned with the on-screen speaker, as audio originating from off-screen individuals or background music can disrupt the modeling of audio-visual correspondence and negatively impact detection accuracy. 
    
    \noindent\textbf{Future work.} To address the limitations above, future work will explore three main directions. Firstly, we plan to leverage large language models (LLMs) to develop a more flexible detection framework capable of handling incomplete or single-modal content. Secondly, to mitigate the reliance on aligned audio-visual inputs, we aim to incorporate speaker localization and audio source separation techniques. Finally, we will attempt to delve into finer-grained audio features (e.g., phonemes) and utilize LLMs to detect temporal inconsistencies between phonemes and facial action units (FAUs), thereby enhancing generalization capabilities and interpretability of forgery detectors.


 





\bibliographystyle{IEEEtran}
\bibliography{ref}

\begin{thebibliography}{10}
\providecommand{\url}[1]{#1}
\csname url@samestyle\endcsname
\providecommand{\newblock}{\relax}
\providecommand{\bibinfo}[2]{#2}
\providecommand{\BIBentrySTDinterwordspacing}{\spaceskip=0pt\relax}
\providecommand{\BIBentryALTinterwordstretchfactor}{4}
\providecommand{\BIBentryALTinterwordspacing}{\spaceskip=\fontdimen2\font plus
\BIBentryALTinterwordstretchfactor\fontdimen3\font minus \fontdimen4\font\relax}
\providecommand{\BIBforeignlanguage}[2]{{%
\expandafter\ifx\csname l@#1\endcsname\relax
\typeout{** WARNING: IEEEtran.bst: No hyphenation pattern has been}%
\typeout{** loaded for the language `#1'. Using the pattern for}%
\typeout{** the default language instead.}%
\else
\language=\csname l@#1\endcsname
\fi
#2}}
\providecommand{\BIBdecl}{\relax}
\BIBdecl

\bibitem{blasberg2023you}
J.~U. Blasberg, M.~Gallistl, M.~Degering, F.~Baierlein, and V.~Engert, ``You look stressed: A pilot study on facial action unit activity in the context of psychosocial stress,'' \emph{Comprehensive Psychoneuroendocrinology}, vol.~15, p. 100187, 2023.

\bibitem{chugh2020not}
K.~Chugh, P.~Gupta, A.~Dhall, and R.~Subramanian, ``Not made for each other-audio-visual dissonance-based deepfake detection and localization,'' in \emph{Proceedings of the ACM International Conference on Multimedia}, 2020, pp. 439--447.

\bibitem{hashmi2022multimodal}
A.~Hashmi, S.~A. Shahzad, W.~Ahmad, C.~W. Lin, Y.~Tsao, and H.-M. Wang, ``Multimodal forgery detection using ensemble learning,'' in \emph{Asia-Pacific Signal and Information Processing Association Annual Summit and Conference}.\hskip 1em plus 0.5em minus 0.4em\relax IEEE, 2022, pp. 1524--1532.

\bibitem{hashmi2025avtenet}
A.~Hashmi, S.~A. Shahzad, C.-W. Lin, Y.~Tsao, and H.-M. Wang, ``Avtenet: A human-cognition-inspired audio-visual transformer-based ensemble network for video deepfake detection,'' \emph{IEEE Transactions on Cognitive and Developmental Systems}, 2025.

\bibitem{yu2023pvass}
Y.~Yu, X.~Liu, R.~Ni, S.~Yang, Y.~Zhao, and A.~C. Kot, ``Pvass-mdd: predictive visual-audio alignment self-supervision for multimodal deepfake detection,'' \emph{IEEE Transactions on Circuits and Systems for Video Technology}, 2023.

\bibitem{liu2023mcl}
X.~Liu, Y.~Yu, X.~Li, and Y.~Zhao, ``Mcl: multimodal contrastive learning for deepfake detection,'' \emph{IEEE Transactions on Circuits and Systems for Video Technology}, 2023.

\bibitem{zou2024cross}
H.~Zou, M.~Shen, Y.~Hu, C.~Chen, E.~S. Chng, and D.~Rajan, ``Cross-modality and within-modality regularization for audio-visual deepfake detection,'' in \emph{IEEE International Conference on Acoustics, Speech and Signal Processing}.\hskip 1em plus 0.5em minus 0.4em\relax IEEE, 2024, pp. 4900--4904.

\bibitem{oorloff2024avff}
T.~Oorloff, S.~Koppisetti, N.~Bonettini, D.~Solanki, B.~Colman, Y.~Yacoob, A.~Shahriyari, and G.~Bharaj, ``Avff: Audio-visual feature fusion for video deepfake detection,'' in \emph{Proceedings of the IEEE/CVF Conference on Computer Vision and Pattern Recognition}, 2024, pp. 27\,102--27\,112.

\bibitem{zhang2024joint}
Y.~Zhang, W.~Lin, and J.~Xu, ``Joint audio-visual attention with contrastive learning for more general deepfake detection,'' \emph{ACM Transactions on Multimedia Computing, Communications and Applications}, vol.~20, no.~5, pp. 1--23, 2024.

\bibitem{yin2024fine}
Q.~Yin, W.~Lu, X.~Cao, X.~Luo, Y.~Zhou, and J.~Huang, ``Fine-grained multimodal deepfake classification via heterogeneous graphs,'' \emph{International Journal of Computer Vision}, pp. 1--15, 2024.

\bibitem{mittal2020emotions}
T.~Mittal, U.~Bhattacharya, R.~Chandra, A.~Bera, and D.~Manocha, ``Emotions don't lie: An audio-visual deepfake detection method using affective cues,'' in \emph{Proceedings of the ACM International Conference on Multimedia}, 2020, pp. 2823--2832.

\bibitem{li2020face}
L.~Li, J.~Bao, T.~Zhang, H.~Yang, D.~Chen, F.~Wen, and B.~Guo, ``Face x-ray for more general face forgery detection,'' in \emph{Proceedings of the IEEE/CVF Conference on Computer Vision and Pattern Recognition}, 2020, pp. 5001--5010.

\bibitem{kong2022detect}
C.~Kong, B.~Chen, H.~Li, S.~Wang, A.~Rocha, and S.~Kwong, ``Detect and locate: Exposing face manipulation by semantic-and noise-level telltales,'' \emph{IEEE Transactions on Information Forensics and Security}, vol.~17, pp. 1741--1756, 2022.

\bibitem{wang2022lisiam}
J.~Wang, Y.~Sun, and J.~Tang, ``Lisiam: Localization invariance siamese network for deepfake detection,'' \emph{IEEE Transactions on Information Forensics and Security}, vol.~17, pp. 2425--2436, 2022.

\bibitem{yang2023masked}
Z.~Yang, J.~Liang, Y.~Xu, X.-Y. Zhang, and R.~He, ``Masked relation learning for deepfake detection,'' \emph{IEEE Transactions on Information Forensics and Security}, vol.~18, pp. 1696--1708, 2023.

\bibitem{yan2023ucf}
Z.~Yan, Y.~Zhang, Y.~Fan, and B.~Wu, ``Ucf: Uncovering common features for generalizable deepfake detection,'' in \emph{Proceedings of the IEEE/CVF International Conference on Computer Vision}, 2023, pp. 22\,412--22\,423.

\bibitem{zhang2024robust}
K.~Zhang, Z.~Hua, Y.~Zhang, Y.~Guo, and T.~Xiang, ``Robust ai-synthesized speech detection using feature decomposition learning and synthesizer feature augmentation,'' \emph{IEEE Transactions on Information Forensics and Security}, 2024.

\bibitem{yang2020preventing}
C.-Z. Yang, J.~Ma, S.~Wang, and A.~W.-C. Liew, ``Preventing deepfake attacks on speaker authentication by dynamic lip movement analysis,'' \emph{IEEE Transactions on Information Forensics and Security}, vol.~16, pp. 1841--1854, 2020.

\bibitem{mukhopadhyay2024diff2lip}
S.~Mukhopadhyay, S.~Suri, R.~T. Gadde, and A.~Shrivastava, ``Diff2lip: Audio conditioned diffusion models for lip-synchronization,'' in \emph{Proceedings of the IEEE/CVF Winter Conference on Applications of Computer Vision}, 2024, pp. 5292--5302.

\bibitem{li2024latentsync}
C.~Li, C.~Zhang, W.~Xu, J.~Lin, J.~Xie, W.~Feng, B.~Peng, C.~Chen, and W.~Xing, ``Latentsync: Taming audio-conditioned latent diffusion models for lip sync with syncnet supervision,'' \emph{arXiv preprint arXiv:2412.09262}, 2024.

\bibitem{jiang2024loopy}
J.~Jiang, C.~Liang, J.~Yang, G.~Lin, T.~Zhong, and Y.~Zheng, ``Loopy: Taming audio-driven portrait avatar with long-term motion dependency,'' in \emph{The International Conference on Learning Representations}, 2025.

\bibitem{zhou2021joint}
Y.~Zhou and S.-N. Lim, ``Joint audio-visual deepfake detection,'' in \emph{Proceedings of the IEEE/CVF International Conference on Computer Vision}, 2021, pp. 14\,800--14\,809.

\bibitem{shahzad2022lip}
S.~A. Shahzad, A.~Hashmi, S.~Khan, Y.-T. Peng, Y.~Tsao, and H.-M. Wang, ``Lip sync matters: A novel multimodal forgery detector,'' in \emph{Asia-Pacific Signal and Information Processing Association Annual Summit and Conference}.\hskip 1em plus 0.5em minus 0.4em\relax IEEE, 2022, pp. 1885--1892.

\bibitem{knafo2022fakeout}
G.~Knafo and O.~Fried, ``Fakeout: Leveraging out-of-domain self-supervision for multi-modal video deepfake detection,'' \emph{arXiv preprint arXiv:2212.00773}, 2022.

\bibitem{cozzolino2023audio}
D.~Cozzolino, A.~Pianese, M.~Nie{\ss}ner, and L.~Verdoliva, ``Audio-visual person-of-interest deepfake detection,'' in \emph{Proceedings of the IEEE/CVF Conference on Computer Vision and Pattern Recognition}, 2023, pp. 943--952.

\bibitem{ilyas2023avfakenet}
H.~Ilyas, A.~Javed, and K.~M. Malik, ``Avfakenet: A unified end-to-end dense swin transformer deep learning model for audio--visual deepfakes detection,'' \emph{Applied Soft Computing}, vol. 136, p. 110124, 2023.

\bibitem{raza2023multimodaltrace}
M.~A. Raza and K.~M. Malik, ``Multimodaltrace: Deepfake detection using audiovisual representation learning,'' in \emph{Proceedings of the IEEE/CVF Conference on Computer Vision and Pattern Recognition (CVPR) Workshops}, June 2023, pp. 993--1000.

\bibitem{agarwal2020detecting}
S.~Agarwal, H.~Farid, O.~Fried, and M.~Agrawala, ``Detecting deep-fake videos from phoneme-viseme mismatches,'' in \emph{Proceedings of the IEEE/CVF Conference on Computer Vision and Pattern Recognition Workshops}, 2020, pp. 660--661.

\bibitem{hal2022l}
A.~Haliassos, R.~Mira, S.~Petridis, and M.~Pantic, ``Leveraging real talking faces via self-supervision for robust forgery detection,'' in \emph{Proceedings of the IEEE/CVF Conference on Computer Vision and Pattern Recognition}, 2022, pp. 14\,950--14\,962.

\bibitem{zhao2024audio}
H.~Zhao, W.~Zhou, D.~Chen, W.~Zhang, Y.~Guo, Z.~Cheng, P.~Yan, and N.~Yu, ``Audio-visual contrastive pre-train for face forgery detection,'' \emph{ACM Transactions on Multimedia Computing, Communications and Applications}, 2024.

\bibitem{liu2024lips}
W.~Liu, T.~She, J.~Liu, B.~Li, D.~Yao, and R.~Wang, ``Lips are lying: Spotting the temporal inconsistency between audio and visual in lip-syncing deepfakes,'' in \emph{Advances in Neural Information Processing Systems}, vol.~37, 2024.

\bibitem{peng2024deepfakes}
C.~Peng, Z.~Miao, D.~Liu, N.~Wang, R.~Hu, and X.~Gao, ``Where deepfakes gaze at? spatial-temporal gaze inconsistency analysis for video face forgery detection,'' \emph{IEEE Transactions on Information Forensics and Security}, 2024.

\bibitem{chu2022protecting}
B.~Chu, W.~You, Z.~Yang, L.~Zhou, and R.~Wang, ``Protecting world leader using facial speaking pattern against deepfakes,'' \emph{IEEE Signal Processing Letters}, vol.~29, pp. 2078--2082, 2022.

\bibitem{bai2023aunet}
W.~Bai, Y.~Liu, Z.~Zhang, B.~Li, and W.~Hu, ``Aunet: Learning relations between action units for face forgery detection,'' in \emph{Proceedings of the IEEE/CVF Conference on Computer Vision and Pattern Recognition}, 2023, pp. 24\,709--24\,719.

\bibitem{meng2017listen}
Z.~Meng, S.~Han, and Y.~Tong, ``Listen to your face: Inferring facial action units from audio channel,'' \emph{IEEE Transactions on Affective Computing}, vol.~10, no.~4, pp. 537--551, 2017.

\bibitem{h2021lips}
A.~Haliassos, K.~Vougioukas, S.~Petridis, and M.~Pantic, ``Lips don't lie: A generalisable and robust approach to face forgery detection,'' in \emph{Proceedings of the IEEE/CVF Conference on Computer Vision and Pattern Recognition}, 2021, pp. 5039--5049.

\bibitem{gu2021spatiotemporal}
Z.~Gu, Y.~Chen, T.~Yao, S.~Ding, J.~Li, F.~Huang, and L.~Ma, ``Spatiotemporal inconsistency learning for deepfake video detection,'' in \emph{Proceedings of the ACM International Conference on Multimedia}, 2021, pp. 3473--3481.

\bibitem{zhao2023istvt}
C.~Zhao, C.~Wang, G.~Hu, H.~Chen, C.~Liu, and J.~Tang, ``Istvt: interpretable spatial-temporal video transformer for deepfake detection,'' \emph{IEEE Transactions on Information Forensics and Security}, vol.~18, pp. 1335--1348, 2023.

\bibitem{coccomini2024mintime}
D.~A. Coccomini, G.~K. Zilos, G.~Amato, R.~Caldelli, F.~Falchi, S.~Papadopoulos, and C.~Gennaro, ``Mintime: multi-identity size-invariant video deepfake detection,'' \emph{IEEE Transactions on Information Forensics and Security}, 2024.

\bibitem{gu2021deepfake}
Y.~Gu, X.~Zhao, C.~Gong, and X.~Yi, ``Deepfake video detection using audio-visual consistency,'' in \emph{International Workshop on Digital Watermarking}.\hskip 1em plus 0.5em minus 0.4em\relax Springer, 2021, pp. 168--180.

\bibitem{feng2023self}
C.~Feng, Z.~Chen, and A.~Owens, ``Self-supervised video forensics by audio-visual anomaly detection,'' in \emph{Proceedings of the IEEE/CVF Conference on Computer Vision and Pattern Recognition}, 2023, pp. 10\,491--10\,503.

\bibitem{cheng2023voice}
H.~Cheng, Y.~Guo, T.~Wang, Q.~Li, X.~Chang, and L.~Nie, ``Voice-face homogeneity tells deepfake,'' \emph{ACM Transactions on Multimedia Computing, Communications and Applications}, vol.~20, no.~3, pp. 1--22, 2023.

\bibitem{bohacek2024lost}
M.~Bohacek and H.~Farid, ``Lost in translation: Lip-sync deepfake detection from audio-video mismatch,'' in \emph{Proceedings of the IEEE/CVF Conference on Computer Vision and Pattern Recognition}, 2024, pp. 4315--4323.

\bibitem{yang2023avoid}
W.~Yang, X.~Zhou, Z.~Chen, B.~Guo, Z.~Ba, Z.~Xia, X.~Cao, and K.~Ren, ``Avoid-df: Audio-visual joint learning for detecting deepfake,'' \emph{IEEE Transactions on Information Forensics and Security}, vol.~18, pp. 2015--2029, 2023.

\bibitem{liangspeechforensics}
Y.~Liang, M.~Yu, G.~Li, J.~Jiang, B.~Li, F.~Yu, N.~Zhang, X.~Meng, and W.~Huang, ``Speechforensics: Audio-visual speech representation learning for face forgery detection,'' in \emph{Advances in Neural Information Processing Systems}, 2024.

\bibitem{chen2025glcf}
X.~Chen, Q.~Yin, J.~Liu, W.~Lu, X.~Luo, and J.~Zhou, ``Glcf: A global-local multimodal coherence analysis framework for talking face generation detection,'' in \emph{Proceedings of the AAAI Conference on Artificial Intelligence}, vol.~39, no.~1, 2025, pp. 75--83.

\bibitem{vaswani2017attention}
A.~Vaswani, N.~Shazeer, N.~Parmar, J.~Uszkoreit, L.~Jones, A.~N. Gomez, {\L}.~Kaiser, and I.~Polosukhin, ``Attention is all you need,'' \emph{Advances in Neural Information Processing Systems}, vol.~30, 2017.

\bibitem{li2023blip}
J.~Li, D.~Li, S.~Savarese, and S.~Hoi, ``Blip-2: Bootstrapping language-image pre-training with frozen image encoders and large language models,'' in \emph{International Conference on Machine Learning}.\hskip 1em plus 0.5em minus 0.4em\relax PMLR, 2023, pp. 19\,730--19\,742.

\bibitem{tran2019video}
D.~Tran, H.~Wang, L.~Torresani, and M.~Feiszli, ``Video classification with channel-separated convolutional networks,'' in \emph{Proceedings of the IEEE/CVF International Conference on Computer Vision}, 2019, pp. 5552--5561.

\bibitem{LuoS0SG22}
C.~Luo, S.~Song, W.~Xie, L.~Shen, and H.~Gunes, ``Learning multi-dimensional edge feature-based {AU} relation graph for facial action unit recognition,'' in \emph{International Joint Conference on Artificial Intelligence}, 2022, pp. 1239--1246.

\bibitem{radford2023robust}
A.~Radford, J.~W. Kim, T.~Xu, G.~Brockman, C.~McLeavey, and I.~Sutskever, ``Robust speech recognition via large-scale weak supervision,'' in \emph{International Conference on Machine Learning}.\hskip 1em plus 0.5em minus 0.4em\relax PMLR, 2023, pp. 28\,492--28\,518.

\bibitem{mavadati2013disfa}
S.~M. Mavadati, M.~H. Mahoor, K.~Bartlett, P.~Trinh, and J.~F. Cohn, ``Disfa: A spontaneous facial action intensity database,'' \emph{IEEE Transactions on Affective Computing}, vol.~4, no.~2, pp. 151--160, 2013.

\bibitem{cai2022you}
Z.~Cai, K.~Stefanov, A.~Dhall, and M.~Hayat, ``Do you really mean that? content driven audio-visual deepfake dataset and multimodal method for temporal forgery localization,'' in \emph{International Conference on Digital Image Computing: Techniques and Applications}.\hskip 1em plus 0.5em minus 0.4em\relax IEEE, 2022, pp. 1--10.

\bibitem{khalid2021fakeavceleb}
H.~Khalid, S.~Tariq, M.~Kim, and S.~S. Woo, ``Fakeavceleb: A novel audio-video multimodal deepfake dataset,'' \emph{arXiv preprint arXiv:2108.05080}, 2021.

\bibitem{tak2021end}
H.~Tak, J.-W. Jung, J.~Patino, M.~Kamble, M.~Todisco, and N.~Evans, ``End-to-end spectro-temporal graph attention networks for speaker verification anti-spoofing and speech deepfake detection,'' in \emph{ASVSPOOF, Automatic Speaker Verification and Spoofing Countermeasures Challenge}.\hskip 1em plus 0.5em minus 0.4em\relax ISCA, 2021, pp. 1--8.

\bibitem{tak2022automatic}
H.~Tak, M.~Todisco, X.~Wang, J.-w. Jung, J.~Yamagishi, and N.~Evans, ``Automatic speaker verification spoofing and deepfake detection using wav2vec 2.0 and data augmentation,'' in \emph{The Speaker and Language Recognition Workshop}.\hskip 1em plus 0.5em minus 0.4em\relax ISCA, 2022.

\bibitem{sung2023hearing}
C.-S. Sung, J.-C. Chen, and C.-S. Chen, ``Hearing and seeing abnormality: Self-supervised audio-visual mutual learning for deepfake detection,'' in \emph{IEEE International Conference on Acoustics, Speech and Signal Processing}.\hskip 1em plus 0.5em minus 0.4em\relax IEEE, 2023, pp. 1--5.

\bibitem{heo2020adamp}
B.~Heo, S.~Chun, S.~J. Oh, D.~Han, S.~Yun, G.~Kim, Y.~Uh, and J.-W. Ha, ``Adamp: Slowing down the slowdown for momentum optimizers on scale-invariant weights,'' \emph{arXiv preprint arXiv:2006.08217}, 2020.

\bibitem{jiang2020deeperforensics}
L.~Jiang, R.~Li, W.~Wu, C.~Qian, and C.~C. Loy, ``Deeperforensics-1.0: A large-scale dataset for real-world face forgery detection,'' in \emph{Proceedings of the IEEE/CVF Conference on Computer Vision and Pattern Recognition}, 2020, pp. 2889--2898.

\bibitem{van2008visualizing}
L.~Van~der Maaten and G.~Hinton, ``Visualizing data using t-sne.'' \emph{Journal of Machine Learning Research}, vol.~9, no.~11, 2008.

\end{thebibliography}

\vfill

\begin{IEEEbiography}[{\includegraphics[width=1in,height=1.25in,clip,keepaspectratio]{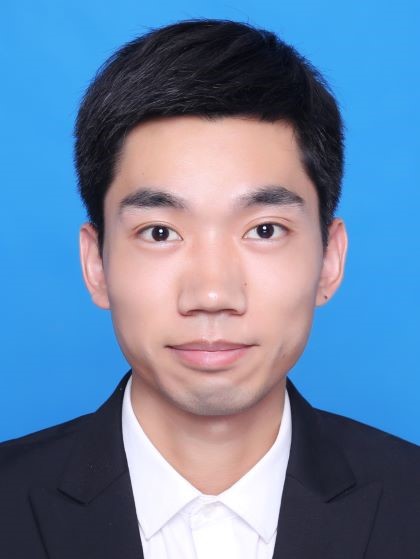}}]{Jian Wang}
received the Ph.D. degree at Intelligent Media Analysis Group (IMAG), Nanjing University of Science and Technology, China. He is currently a postdoctoral researcher at the School of Computer Science, Nanjing University of Posts and Telecommunications, China. His research interests include multimedia forensics and computer vision. 
\end{IEEEbiography}

\begin{IEEEbiography}[{\includegraphics
[width=0.8in,height=1.25in,clip,
keepaspectratio]{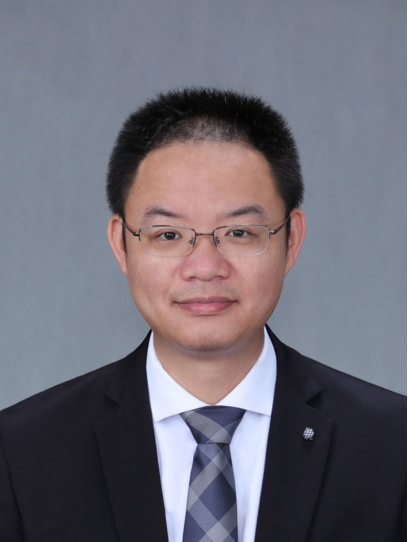}}]
{Baoyuan Wu} (Senior Member, IEEE) is a Tenured Associate Professor of the School of Data Science, Chinese University of Hong Kong, Shenzhen, Guangdong, 518172, P.R. China. His research interests are trustworthy and generative AI. He has published 100+ top-tier conference and journal papers. He is currently serving as an Associate Editor of IEEE Transactions on Information Forensics and Security and Neurocomputing, Organizing Chair of PRCV 2022, and Area Chairs of several top-tier conferences. He received the ``2023 Young Researcher Award” of The Chinese University of Hong Kong, Shenzhen.
\end{IEEEbiography}

\begin{IEEEbiography}[{\includegraphics
[width=0.8in,height=1.25in,clip,
keepaspectratio]{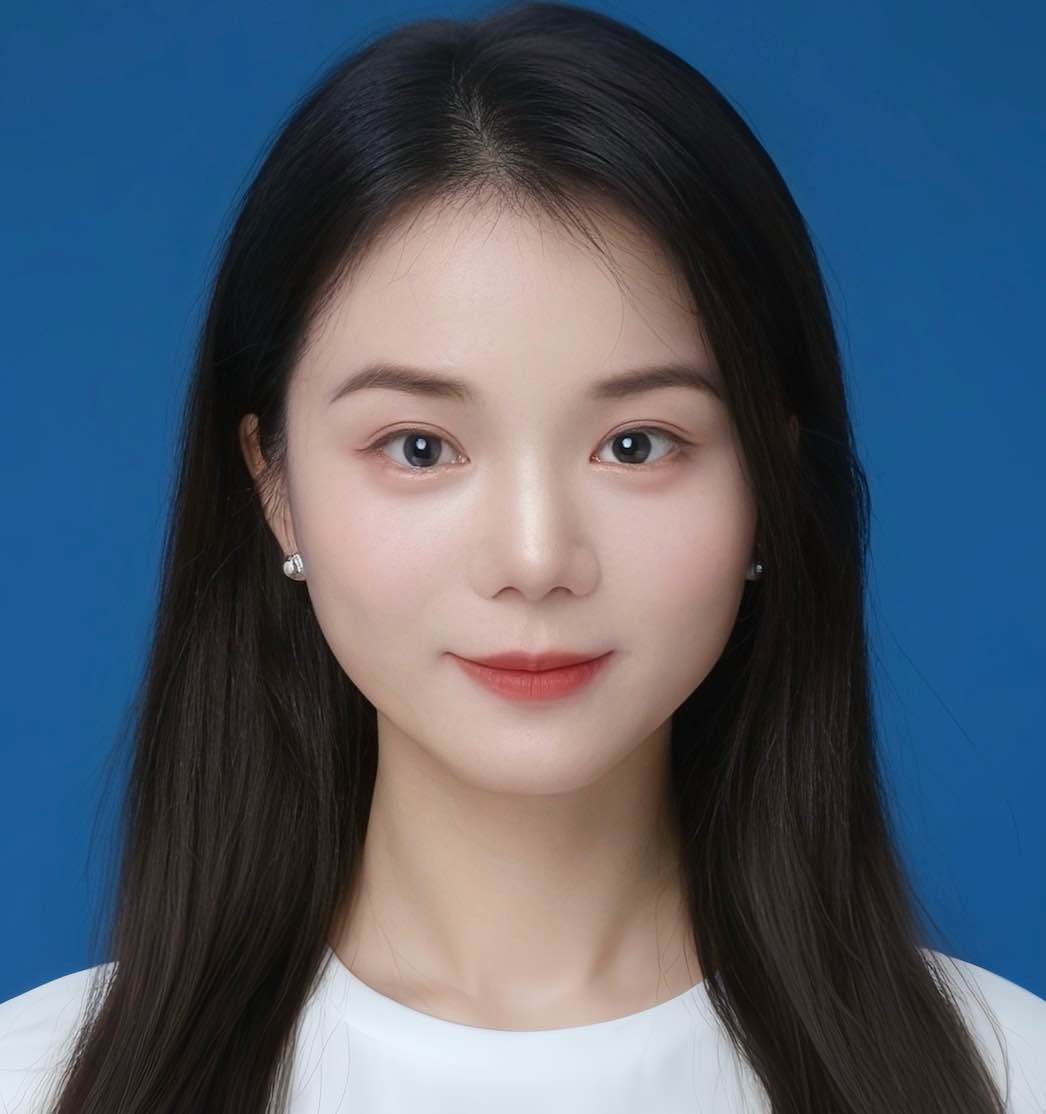}}]
{Li Liu} (Senior Member, IEEE) is currently an Assistant Professor at the Artificial Intelligence Thrust, Information Hub, Hong Kong University of Science and Technology (Guangzhou). She received her Ph.D. degree from Gipsa-lab, University Grenoble Alpes, France. Her research focuses on audio-visual speech processing and Trustworthy AI. She has published over 60 papers in top-tier journals and conferences, including IEEE TPAMI, IEEE TASLP, IEEE TMM, NeurIPS, IJCAI, ACM MM, ICASSP etc. Dr. Liu is an IEEE senior member and she is an active member of the IEEE SPS, serving on the IEEE Machine Learning for Signal Processing Technical Committee (MLSP-TC) and as Chair of the election subcommittee. She has also contributed to several IEEE conferences, including serving as the China-site Local Chair for ICASSP 2022 and Area Chair ICASSP 2023, 2024. She won the French Sephora Berribi Award for Female Scientists in Mathematics and Computers, and obtained Outstanding Contribution Award (2023, 2024) in Hong Kong University of Science and Technology (Guangzhou).
\end{IEEEbiography}

\begin{IEEEbiography}[{\includegraphics[width=1in,height=1.25in,clip,keepaspectratio]{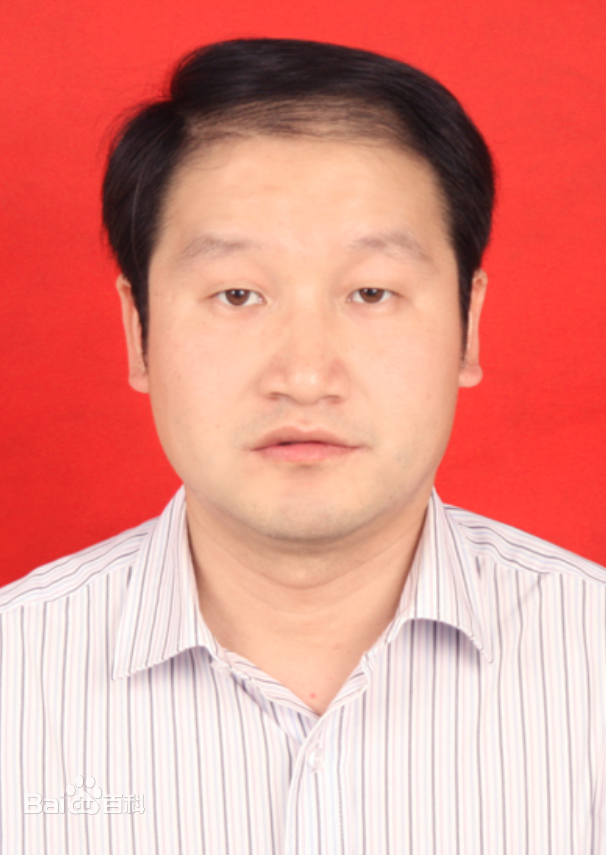}}]{Qingshan Liu} (Senior Member, IEEE) received the master degree from Southeast University, Nanjing, China, in 2000 and the PhD degree from the Chinese Academy of Sciences, Beijing, China, in 2003. From 2010 to 2011, he was an assistant research professor in the Department of Computer Science, Computational Biomedicine Imaging and Modeling Center, Rutgers, State University of New Jersey, Piscataway, New Jersey. Before he joined Rutgers University, he was an associate professor in the National Laboratory of Pattern Recognition, Chinese Academy of Sciences. He is currently a professor in Nanjing University of Posts and Telecommunications and PhD supervisor of Nanjing University of Aeronautics and Astronautics. His research interests include image and vision analysis, machine learning, etc.
\end{IEEEbiography}

\end{document}